\documentclass[10pt,twocolumn,letterpaper]{article}

\usepackage{cvpr}              % To produce the CAMERA-READY version
\usepackage{times}
\usepackage{epsfig}
\usepackage{graphicx}
\usepackage{amsmath}
\usepackage{amssymb}
\usepackage{mathtools}
\usepackage{multirow}
\usepackage{enumitem}
\usepackage{caption}
\usepackage{subcaption}
\usepackage{array}
\usepackage{colortbl}
\usepackage{booktabs}
\usepackage{bbm}
\usepackage{multicol}
\usepackage{makecell}
\usepackage{xcolor}
\usepackage{xspace}
\usepackage{float}
\usepackage{color}
\usepackage{siunitx}
\usepackage{pifont}
\usepackage{marvosym}
\usepackage{CJKutf8}
\usepackage{tabularx}

\usepackage[breakable]{tcolorbox}

\definecolor{linkColor}{rgb}{0.18,0.39,0.62}
\definecolor{cvprblue}{rgb}{0.21,0.49,0.74}
\usepackage[pagebackref,breaklinks,colorlinks,allcolors=cvprblue]{hyperref}

\definecolor{baselinecolor}{gray}{.9}

\definecolor{promptBlueBack}{HTML}{EBF5FF}
\definecolor{promptBlueFrame}{HTML}{003366}
\definecolor{promptGreenBack}{HTML}{F0FFF0}
\definecolor{promptGreenFrame}{HTML}{004D00}
\definecolor{promptOrangeBack}{HTML}{FFF5E6}
\definecolor{promptOrangeFrame}{HTML}{D35400} 
\definecolor{mylightgray}{gray}{0.9}

\newcommand\cmark{\textcolor[RGB]{13, 120, 24}{\ding{52}}}

\newcommand\xmark{\textcolor[RGB]{173, 21, 14}{\ding{55}}}

\newcommand{\modelname}{{AndroidLens}\xspace}

\newtcolorbox{bluepromptbox}[1]{
    colback=promptBlueBack,   
    colframe=promptBlueFrame, 
    title=#1,                 
    fonttitle=\bfseries, 
    boxrule=0.5mm, 
    arc=2mm, 
    left=2mm,
    right=2mm, 
    top=2mm,
    bottom=2mm,
    breakable,
}

\newtcolorbox{greenpromptbox}[1]{
    colback=promptGreenBack,    
    colframe=promptGreenFrame,  
    title=#1,             
    fonttitle=\bfseries, 
    boxrule=0.5mm, 
    arc=2mm, 
    left=2mm,
    right=2mm, 
    top=2mm,
    bottom=2mm,
    breakable,
}

\newtcolorbox{examplebox}[1]{
    colback=promptOrangeBack,    
    colframe=promptOrangeFrame,  
    title=#1,             
    fonttitle=\bfseries, 
    boxrule=0.5mm, 
    arc=2mm, 
    left=2mm,
    right=2mm, 
    top=2mm,
    bottom=2mm,
    breakable,
}

\def\confName{CVPR}
\def\confYear{2026}

\title{AndroidLens: Long-latency Evaluation with Nested Sub-targets for Android GUI Agents}

\author{First Author\\
Institution1\\
Institution1 address\\
{\tt\small firstauthor@i1.org}
\and
Second Author\\
Institution2\\
First line of institution2 address\\
{\tt\small secondauthor@i2.org}
}

\author{
Yue Cao$^{1,2}$\thanks{Equal contribution, random order.}~~\thanks{This work was done during an internship at Alibaba.}, 
Yingyao Wang$^{2}$\footnotemark[1], 
Pi Bu$^{2}$\footnotemark[1], 
Jingxuan Xing$^{2}$\footnotemark[1], 
Wei Jiang$^{2}$\footnotemark[1],
Zekun Zhu$^{2}$,  \\ 
Junpeng Ma$^{3, 2}$\footnotemark[2], 
Sashuai Zhou$^{4, 2}$\footnotemark[2], 
Tong Lu$^{1}$\thanks{Corresponding Authors.},
Jun Song$^{2}$\footnotemark[2], 
Yu Cheng$^{2}$, 
Yuning Jiang$^{2}$, 
Bo Zheng$^{2}$\\
$^1$Nanjing University\;$^2$Alibaba Group\;$^3$Fudan University\; $^4$Zhejiang University \\
{\small Code: \url{https://github.com/alibaba/AndroidLens}} \\
{\small Data: \url{https://huggingface.co/datasets/yuecao0119/AndroidLens}} \\
}

\begin{document}

\maketitle

\begin{abstract}

Graphical user interface (GUI) agents can substantially improve productivity by automating frequently executed long-latency tasks on mobile devices. However, existing evaluation benchmarks are still constrained to limited applications, simple tasks, and coarse-grained metrics.
To address this, we introduce \modelname, a challenging evaluation framework for mobile GUI agents, comprising 571 long-latency tasks in both Chinese and English environments, each requiring an average of more than 26 steps to complete.
The framework features: (1) tasks derived from real-world user scenarios across 38 domains, covering complex types such as multi-constraint, multi-goal, and domain-specific tasks; (2) static evaluation that preserves real-world anomalies and allows multiple valid paths to reduce bias; and (3) dynamic evaluation that employs a milestone-based scheme for fine-grained progress measurement via Average Task Progress (ATP).
Our evaluation indicates that even the best models reach only a 12.7\% task success rate and 50.47\% ATP. We also underscore key challenges in real-world environments, including environmental anomalies, adaptive exploration, and long-term memory retention.
% Data and code will be released.
\end{abstract}    
\section{Introduction}
\label{sec:intro}

\begin{table}[!t]
    \centering
    \footnotesize
    \renewcommand\arraystretch{1.1}
    \setlength{\tabcolsep}{2.7pt}
    \caption{\textbf{Comparison of \modelname and other Mobile GUI Agent benchmarks.} \textmd{Column definitions: \# Step (average steps per task), Env. (supports environment interactions), LL (has low-level instructions), GT (provides ground truth trajectories).}}
    \vspace{-2pt}
    \begin{tabular}{p{2cm}ccc>{\centering\arraybackslash}p{0.9cm}>{\centering\arraybackslash}p{0.45cm}>{\centering\arraybackslash}p{0.45cm}>{\centering\arraybackslash}p{0.45cm}}
    \toprule
    \textbf{Dataset} & \textbf{Lang.} & \textbf{\# Tasks} & \textbf{\# Apps} & \textbf{\# Step} & \textbf{Env.} & \textbf{LL} & \textbf{GT} \\
    \midrule
    AITW~\cite{rawles2023androidinthewild} & en &  30,378 & 357 & 6.5 & \xmark & \xmark & \cmark  \\
    AITZ~\cite{zhang2024android} & en &  2,504 & 70 & 7.5 & \xmark & \cmark & \cmark  \\
    GUIOdyssey~\cite{lu2024guiodyssey} & en & 8,334 & 212 & 15.3 & \xmark & \xmark & \xmark  \\
    AndroidControl~\cite{li2024effects} & en &  15,283 & 833 & 4.8 & \xmark  & \cmark & \cmark \\
    AMEX~\cite{chai2024amex} & en &  2,946 & 110 & 12.8 & \xmark & \xmark & \cmark  \\
    AppAgent~\cite{zhang2023appagent} & en &  50 & 10 & - & \xmark  & \xmark & \xmark \\
    \midrule
    LlamaTouch~\cite{zhang2024llamatouch} & en &  496 & 57 & 7.01 & \cmark & \xmark & \cmark  \\
    AndroidWorld~\cite{rawles2024androidworld} & en &  116 & 20 & - & \cmark & \xmark & \xmark \\
    AndroidLab~\cite{xu2024androidlab} & en &  138 & 9 & 8.5 & \cmark & \xmark & \xmark  \\
    LearnGUI~\cite{liu2025learnact} & en &  {2,353} & {73} & {13.2} & {\cmark}  & {\cmark} & {\cmark}  \\
    SPA-Bench~\cite{wang2024mobileagentbench} & en\&zh &  201 & 21 & 8.2 & \cmark & \xmark & \xmark  \\
    \midrule
    \textbf{\modelname} & en\&zh &  \textbf{571} & \textbf{74} & \textbf{26.1} & \textbf{\cmark} & \textbf{\cmark} & \textbf{\cmark}  \\
    \bottomrule
    \end{tabular}
    \vspace{-0.5cm}
    \label{tab:comp_other}
\end{table}

\begin{figure*}
  \centering
    % \fbox{\rule{0pt}{2in} \rule{.9\linewidth}{0pt}}
    \includegraphics[width=\textwidth]{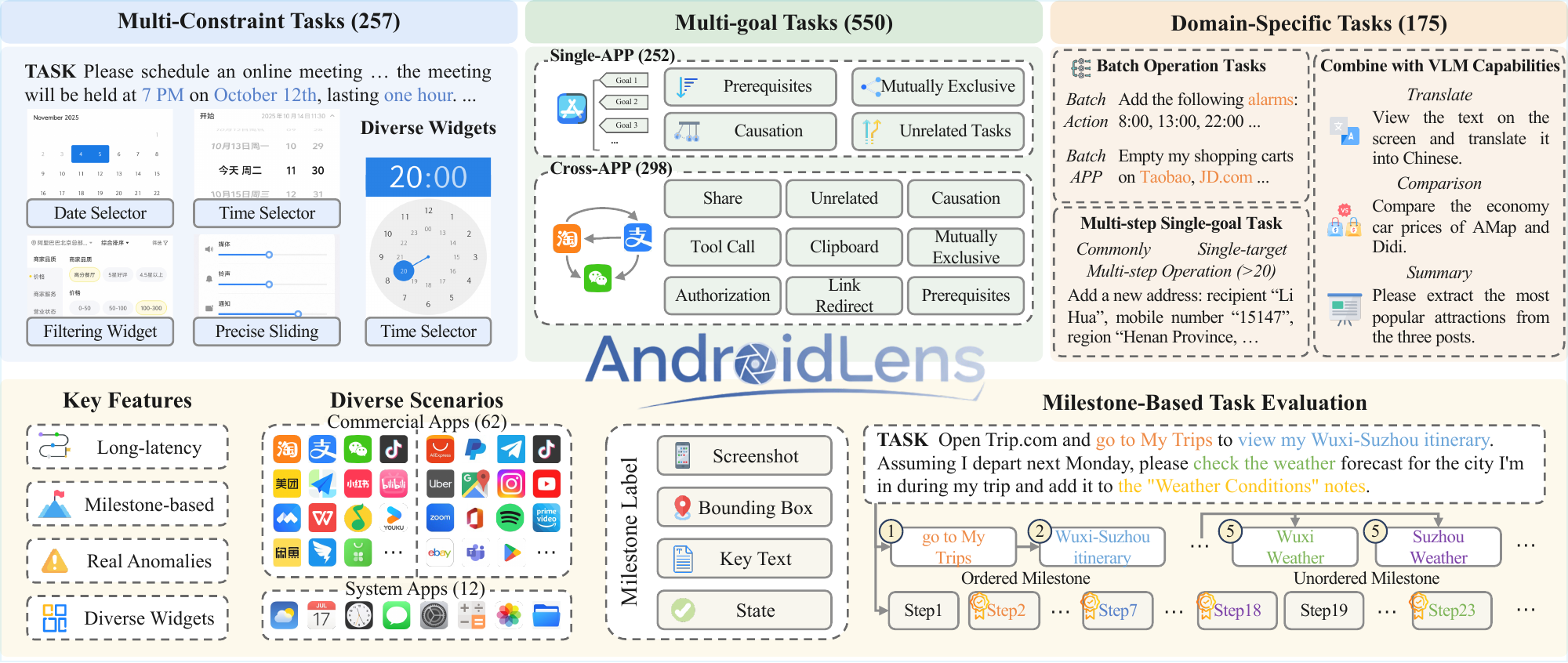}
    \caption{\textbf{An overview of the \modelname benchmark.} Its main characteristics are: (1) It covers long-latency tasks with multiple constraints, multiple objectives, and domain-specific requirements, involving increasingly complex widget operations, cross-app interactions, batch operations, and more. (2) It adopts a milestone-based intermediate-goal evaluation scheme. By using stable and verifiable milestones to measure average task progress, the benchmark provides fine-grained result analysis for long-latency tasks.}
  \label{fig:overview}
\end{figure*}

Enabled by large-scale training corpora, vision–language models (VLMs)~\cite{openai2024gpt4o,bai2025qwen25vl,zhu2025internvl3,team2025kimi,anthropic2025claude37} have achieved remarkable progress in cross-modal perception and instruction following, evolving from passive understanding to interactive decision-making.
These advances have catalyzed the development of graphical user interface (GUI) agents capable of interpreting and manipulating on-screen elements, particularly in mobile environments~\cite{wu2024copilot,cheng2024seeclick,zheng2024seeact,hong2024cogagent}. 
Mobile GUI agents can convert natural language into automated actions across apps without requiring application modification, effectively lowering barriers to digital access and enhancing productivity~\cite{xu2024aguvis,wu2024atlas,lin2024showui,qin2025ui}.

To effectively assess the performance of mobile GUI agents, existing evaluation benchmarks have made notable progress in both static and dynamic evaluation schemes~\cite{cheng2024seeclick,rawles2024androidworld,li2025screenspotpro,chai2024amex, rawles2023androidinthewild, xu2024androidlab}. 
However, several critical limitations remain:
(1) \textit{Limited domain coverage}: Most benchmarks~\cite{rawles2024androidworld,chai2025a3,xu2024androidlab}, especially dynamic benchmarks, primarily target English-language applications, overlooking Chinese applications that often feature richer functionalities and more intricate execution environments.
(2) \textit{Simplistic task design}: Current benchmarks~\cite{rawles2023androidinthewild,zhang2024android,li2024effects} fail to reflect real user scenarios that involve long execution chains, such as multi-page navigation, nested widgets, or abnormal state recovery. Moreover, cross-application tasks are limited to basic shared operations and ignore broader interactive behaviors~\cite{chen2024spa,deng2024mobile}.
(3) \textit{Coarse evaluation granularity}: Static evaluations~\cite{rawles2023androidinthewild,li2024effects,zhang2025agentcpm} typically depend on a single golden trajectory, despite the existence of multiple valid solutions. And dynamic evaluations~\cite{zhang2024llamatouch,rawles2024androidworld,liu2025learnact} often emphasize final outcomes, making it difficult to distinguish agent performance under complex tasks.

To address these issues, we introduce \modelname, a complex instruction evaluation framework for mobile GUI agents that supports static and dynamic evaluation schemes. 
As shown in~\cref{fig:overview}, \modelname comprises 571 complex tasks across 38 everyday domains in both Chinese and English contexts, including 273 single-application tasks and 298 cross-application tasks. 
Through a rigorous annotation and filtering process, \modelname preserves only representative high-quality instructions, with an average of 26 interaction steps per task.
For static evaluation, we intentionally preserve real-world anomalies such as advertisement pop-ups and missing permissions to assess model robustness under unexpected circumstances. 
To ensure fairness, we also provide multiple execution paths of equal length, all converging to the same target state, thereby avoiding bias toward a single trajectory.
For dynamic evaluation, we introduce a milestone-based scheme that reinforces the assessment of intermediate goal completion in long execution chains. We further propose the Average Task Progress (ATP) metric, enabling finer-grained discrimination of model performance during complex task execution, even in cases of incomplete task completion.

Through extensive evaluation and analysis of existing agents~\cite{anthropic2025claude37,openai2025introducing,bai2025qwen25vl,qin2025ui,OS-Atlas-Pro-7B,xu2024aguvis,gu2025ui,lian2025ui} on \modelname, we highlight the following insights:
(1) Agents need to improve their perception of small UI widgets, especially given the complexity and diversity of components in commercial applications. 
They also need finer grained control operations. For example, Swipe actions on time selection components require precise start and end coordinates.
(2) For long-latency tasks, an agent’s memory capability directly impacts its awareness of task progress and thus its overall performance. Frequent screen switching in cross-application tasks further disrupts memory coherence.
(3) Although chain-of-thought models perform better overall, their ability to recover from unexpected states remains limited. In addition, inconsistencies between their reasoning traces and final answers constrain their ultimate performance.

In summary, our comprehensive benchmark makes the following key contributions: 
\begin{itemize}
% [itemsep=2pt,topsep=3pt,parsep=0pt]
\item We introduce a set of long-latency tasks (average 26 steps), covering both Chinese and English, consisting of 571 unique instructions with high-quality, human-annotated expert trajectories.
\item We develop a unified evaluation suite for long-latency tasks, including both static and dynamic schemes, to enable plug-and-play integration with diverse agents. We also propose an average task progress metric that provides more fine-grained insights into model performance.
\item We conduct an extensive empirical study, benchmarking and analyzing 15 agents spanning both agentic workflows and agent-as-a-models.
\end{itemize}

\section{Related Work}
\label{sec:related_work}

\subsection{Mobile GUI Agents}

Mobile GUI agents accomplish tasks within the graphical user interface (GUI) of a mobile device through multimodal perception, task understanding, and action execution.
Current approaches can be broadly categorized into two types:
(1) Agent workflows~\cite{zhang2023appagent, rawles2024androidworld, wang2024mobile2, wang2025mobilee}, which decompose tasks into sub-goals and leverage existing VLMs and architectural designs to realize agent functionality.
M3A~\cite{rawles2024androidworld} employs a fusion of ReAct~\cite{yao2022} and Reflexion~\cite{shinn2023reflexion} prompting strategies.
Mobile-Agent-v2~\cite{wang2024mobile2} introduces a memory module and multi-agent collaboration to improve navigation, while Mobile-Agent-E~\cite{wang2025mobilee} advances a self-evolving hierarchical agent framework.
(2) Agent-as-a-models~\cite{gu2025ui,hong2024cogagent,xu2024aguvis}, which rely on a single model to execute mobile-agent tasks directly, enabling end-to-end and scalable capability expansion.
Recent multimodal general-purpose large models~\cite{bai2025qwen25vl,openai2024gpt4o,zhu2025internvl3} have demonstrated impressive performance on GUI-related tasks. 
Building on this, researchers explores GUI-oriented fine-tuning strategies, including visual enhancement~\cite{hong2024cogagent,OS-Atlas-Pro-7B}, historical trajectory modeling~\cite{lu2025arpo,sun2024genesis}, multi-platform unification~\cite{qin2025ui,xu2024aguvis}, and reinforcement fine-tuning~\cite{gu2025ui,gu2025mobile,zhang2025agentcpm}. 
While these models show competitive results on standard mobile-agent benchmarks~\cite{rawles2024androidworld,li2024effects,rawles2023androidinthewild}, their effectiveness in handling complex, long-latency instructions in real-world contexts remains underexplored.

\subsection{Mobile GUI Benchmarks}

An effective model evaluation reveals model weaknesses and guides improvement.
Researchers~\cite{chai2024amex, rawles2023androidinthewild, zhang2024android, ai2025inquiremobile} design static evaluation schemes based on the golden trajectory to assess step-level accuracy.
AitW~\cite{rawles2023androidinthewild}, AitZ~\cite{zhang2024android}, AndroidControl~\cite{li2024effects}, and GUI-Odyssey~\cite{lu2025guiodyssey} advance static evaluation by building large-scale or finely annotated datasets and benchmarks.
However, these methods depend on predicting actions from a single screenshot and a history of correct actions, which fails to capture the dynamic and interactive nature of real-world environments. 
In practice, one mistake can cascade and severely degrade subsequent performance.
On the other hand, dynamic evaluations~\cite{rawles2024androidworld, xu2024androidlab, wang2024mobileagentbench} assess an agent’s ability to achieve user goals in end-to-end realistic, interactive environments.
AndroidWorld~\cite{rawles2024androidworld} and AndroidLab~\cite{xu2024androidlab} advance the development of dynamic and interactive evaluation benchmarks, but their applications remain limited to offline scenarios. A3~\cite{chai2025a3} further expands the scope of evaluation to include English commercial applications, but mainly focuses on single-app tasks.
In contrast, SPA-Bench~\cite{chen2024spa} proposes an online evaluation framework that also considers Chinese and cross-app interactions. However, its single-app evaluations do not account for complex, multi-subtask instructions, and its cross-app evaluations are restricted to a limited set of interactive operations.

To overcome these limitations, we propose a comprehensive evaluation framework, \modelname, which supports long-latency tasks encompassing commonly used Chinese and English applications. ~\cref{tab:comp_other} compares existing evaluation benchmarks for mobile GUI agents.

\section{\modelname Task}

\begin{figure*}
  \centering
    % \fbox{\rule{0pt}{2in} \rule{.9\linewidth}{0pt}}
    \includegraphics[width=\textwidth]{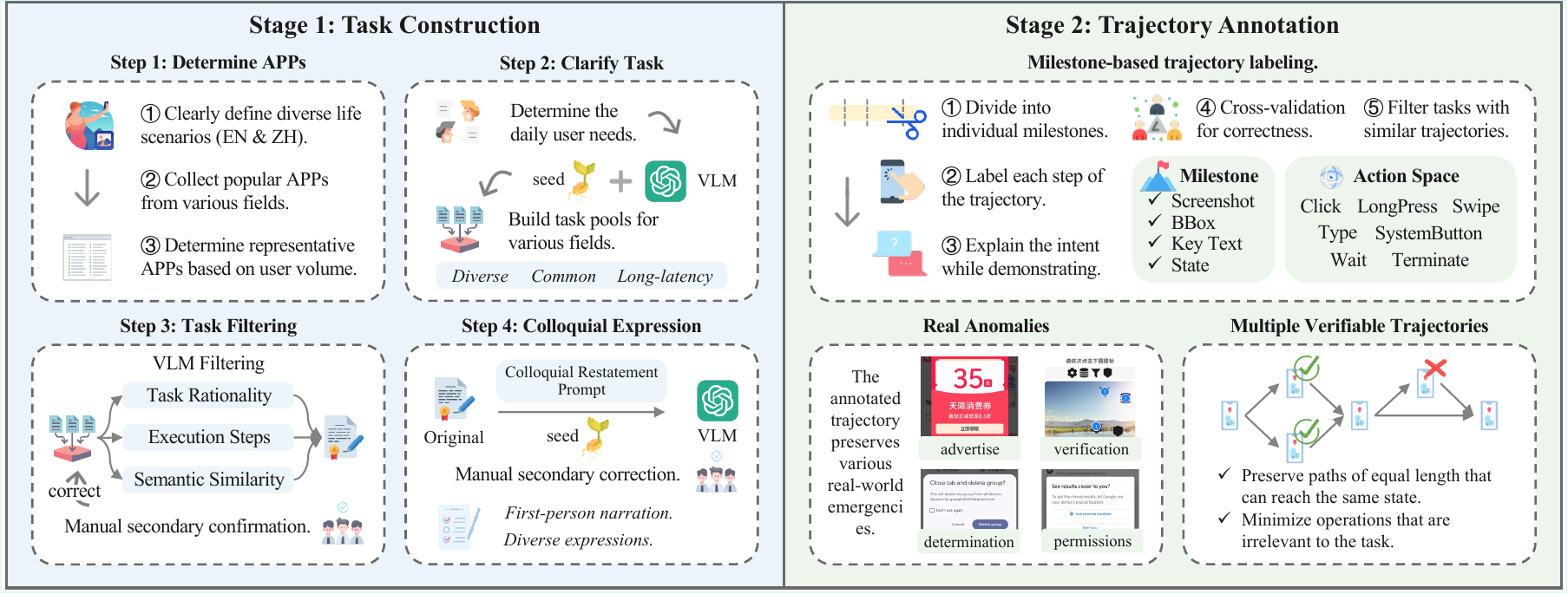}
    \caption{\textbf{Data curation pipeline of \modelname.} The pipeline includes Task Construction, Trajectory Annotation and Quality Control.}
  \label{fig:pipeline}
\end{figure*}

\subsection{Task Definition}
\label{sec:task_def}
A mobile agent is driven by textual instructions, interprets screenshots of mobile screens, and completes task objectives by simulating multi-turn human interactions. Mobile GUI tasks have higher information density and more fine-grained operational requirements, especially in Chinese applications.
We summarize common user interaction habits and define a more general action space, including Click, Swipe, LongPress, Type, System Button, Wait, and Terminate. See Appendix~\ref{sec:action} for the detailed list. Notably, we define Swipe as requiring both a start point and an end point, making it easier to operate common widgets such as time pickers, drop-down lists, and multi-select components.

\subsection{Data Curation Pipeline}
\label{sec:data_pipe}

\paragraph{Task Construction.}

To construct more diverse and representative long-latency tasks, we design a rigorous task-building pipeline. As shown in~\cref{fig:pipeline}, the pipeline includes four stages: 
(1) Determining life scenarios and popular apps: For both Chinese and English environments, we compile 74 apps across 38 daily-life scenarios (see Appendix~\ref{sec:all_app}). Considering the high functional and interface homogeneity among apps within the same scenario, we select only one or two representative apps based on their user scale. (2) Clarifying task: Annotators propose long-latency task requirements that may occur in daily life within these scenarios, including batch operations, multi-step dependent tasks, and more. (3) Task filtering: We deduplicate the collected tasks multiple times to ensure coverage of the major functionalities of the selected apps while keeping the task set diverse. (4) Colloquial expression: We rewrite the finalized tasks into more natural, human-like colloquial expressions rather than structured or formalized commands.

\paragraph{Trajectory Annotation.}

As shown in~\cref{fig:pipeline}, we adopt a human-in-the-loop, step-by-step annotation workflow with multiple verification to ensure annotation quality. The process first splits each task into subtasks based on its verifiability, then performs stepwise execution-style annotation for each subtask while recording the intent behind every step.
To support subsequent subgoal evaluation, we record the final page of each subgoal along with its key page elements as milestones. When selecting milestones, we focus on information that is highly relevant to the task and unlikely to be affected by later page changes. In addition, we document the text, state, and screenshots of key elements in detail to minimize evaluation errors and ensure consistency.

To ensure the reliability of manual annotations, we aim to minimize actions unrelated to the target task. At the same time, we recommend:
(1) retain special cases encountered during annotation (e.g., ad pop-ups, missing permissions, logged-out accounts) to evaluate an agent’s robustness to unexpected events;
(2) keep multiple paths with the same step length, since real tasks often allow several ways to reach the same state. Given annotation costs, we only consider alternative paths of up to three steps.

\paragraph{Quality Control.}
We ensure data quality through a multi-stage data verification process.
(1) Task quality: We examine both the task semantics and the execution trajectories, avoiding not only semantically duplicated tasks but also instructions whose trajectories are highly similar.
(2) Trajectory quality: Annotators cross-checked the execution trajectories of all tasks to ensure the accuracy and simplicity of the trajectories.

\subsection{Dataset Statistics}
\label{sec:data_stat}

\modelname contains 399 Chinese tasks and 172 English tasks, including 273 single-application tasks and 298 cross-application tasks. We follow a rigorous process to ensure task diversity, as Chinese applications typically offer richer functionality and more complex page logic, resulting in a wider variety of tasks.

We conducted a systematic organization and categorization of the tasks. As shown in~\cref{fig:overview}, they can be broadly grouped into three categories:
(1) Multi-constraint tasks require operating according to explicit user preferences or directional goals, and typically involve page widgets such as filters, time, and date selectors.
(2) Multi-goal tasks include multiple related or unrelated subtasks, involving either a single application or multiple applications. In particular, cross-application tasks feature far more diverse interaction patterns than the simple sharing operations found in previous benchmark studies. Detailed subcategories are provided in Appendix~\ref{sec:cross_app}.
(3) Domain-specific tasks include batch operations, tasks that leverage VLM capabilities, and single-goal tasks that require multi-hop reasoning.

We particularly highlight two more challenging types of tasks: 
(1) Batch tasks: These include different operations performed on the same subject, as well as identical operations carried out on different subjects. Automating such batch tasks can significantly improve equipment utilization efficiency.
(2) Tasks leveraging the VLM’s inherent capabilities: These tasks assess whether the agent has forgotten its built-in multimodal abilities, such as translation, information comparison, and summarization.

\begin{figure}[t]
    \centering
    \includegraphics[width=\linewidth]{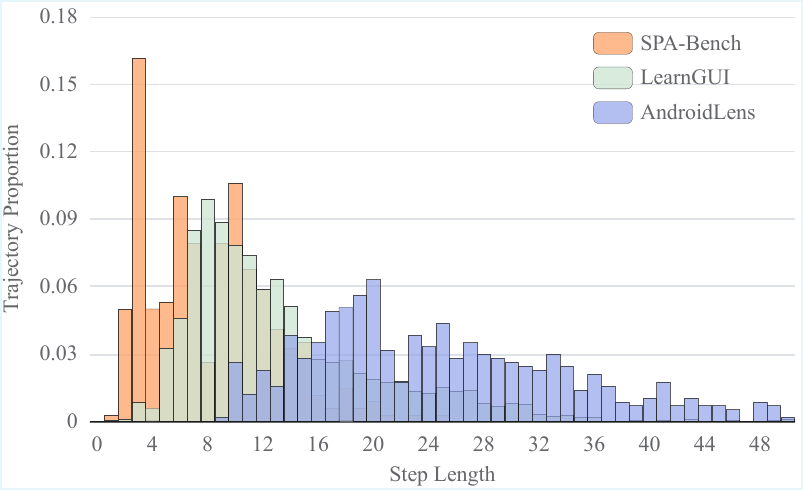}
    % \vspace{-0.4cm}
    \caption{\textbf{Comparison of trajectory lengths between \modelname and existing benchmarks.}}
    \label{fig:traj_length}
    % \vspace{-0.4cm}
\end{figure}

\cref{fig:traj_length} shows that the steps required to complete the \modelname task are significantly more numerous than those in existing benchmarks, highlighting the challenge of our task. In addition, Appendix~\ref{sec:case} presents examples from each category, while Appendix~\ref{sec:statis} provides additional statistical information.

\section{Milestone-Based Task Evaluation}\label{sec:eval_method}

\subsection{Static Evaluation}

For static evaluation, we construct prompts using both high-level (HL) and low-level (LL) instructions for each step in the benchmark and then evaluate the model. High-level instructions provide only a description of the user’s task, whereas low-level instructions describe the expected operation for each individual step. High-level instructions require the model to analyze the current task progress and make decisions, reflecting the model’s historical memory and planning capabilities. In contrast, low-level instructions explicitly state the simple action that should be performed at the current step, directly reflecting the accuracy of the model’s action execution.

\paragraph{Evaluation Metrics.}
At the step level, we use the Action Matching Score (AMS) as the evaluation metric, following GUIOdyssey~\cite{lu2025guiodyssey} and AITW~\cite{rawles2023androidinthewild}. A predicted action type is considered correct only if it exactly matches the ground-truth action type and its attributes. For example, for tap and long-press actions, we check whether the tapped UI element is correct. For swipe actions, we ensure that both the start and end coordinates are correct. It is worth noting that since some models only support providing the start coordinates for swipe actions, we evaluate only whether the swipe direction is correct. For text-input actions, we determine correctness using the average normalized Levenshtein Similarity~\cite{biten2019scene}.
For reasoning agents, we additionally report the number of reasoning output tokens at each step to measure the cost of the model making decisions.

At the task level, we typically consider a task successfully completed only when the entire process is correct, and use this to calculate the Task Success Rate (SR). This metric is relatively strict, and we find that in complex tasks, the results of nearly all agents are close to zero, making it difficult to distinguish model performance. Moreover, long-latency tasks often involve intermediate results that require validation, so supervising only the final outcome is insufficient. Therefore, we propose an Average Task Progress (ATP) metric, which measures the average proportion of milestones an agent completes in each task. This provides a more fine-grained assessment of an agent’s capability by effectively capturing its degree of task completion.

\paragraph{Multi-trajectory Evaluation.} 
In practical applications, the reference relationships between pages are often complex and intertwined, and multiple operation trajectories may lead to the same outcome. As a result, static evaluation schemes that assess only a single trajectory are inherently biased, since different models may exhibit different—but equally valid—reasoning paths. To address this issue, we evaluate multiple operation trajectories of equal length for each intermediate state. A step is considered correct if it aligns with any of these valid trajectories, thereby reducing evaluation errors and improving fairness.

\subsection{Dynamic Evaluation}

For dynamic evaluation, we developed a dedicated evaluation framework to support automated assessment of different agents. In addition, we implemented a hierarchical and fine-grained task-completion evaluation method that leverages milestone information obtained from task annotations.

\paragraph{A Unified Framework.}
To more efficiently integrate different models and fully utilize GPU resources, we designed a decoupled framework that separates agent inference from action execution. The inference module supports both API-based and local model execution, while the execution module directly operates the mobile device via the Android Debug Bridge (ADB). Communication between the two modules is facilitated through a shared directory.
Since most of the applications used in the benchmark are commercial apps, and some of them cannot run reliably in virtualized environments, we prioritize experiments on real devices. To ensure task reliability, we reset the device state using a predefined script before each experimental cycle, thereby providing a stable and consistent environment for evaluation.

\paragraph{Milestone Evaluation.}

For dynamic evaluation, we propose a segmented assessment method that measures the completion status of each milestone.
We categorize the relationships between milestones into two types: 
(1) Ordered matching: Some milestones must be completed in a specific sequence, such as “checking the schedule” followed by “notifying the schedule.”  
(2) Unordered matching: Some consecutive milestones do not require a fixed order and only need to be completed within a given range, such as “filtering by price” and “filtering by brand”.

As shown in \cref{fig:overview}, this process is performed step-by-step based on milestones.
We first split the entire interaction trajectory with screenshots into fixed-length segments (e.g., 10 steps per segment). These are then fed to GPT-4o for judgment. For ordered milestones, starting from the step immediately after the previous milestone is completed, we sequentially use each segment to ask GPT-4o whether the current milestone has been achieved. If it has, GPT-4o returns the index of the “last completion step” for that milestone, which will serve as the starting point for evaluating the next milestone. If not, we move on to the next segment while still evaluating the same milestone.
For unordered milestones, we submit all milestones within the same unordered group to GPT-4o at once and ask it to determine which of them are completed in the given segment, along with the “last completion step” index for each. The detailed workflow is provided in Appendix~\ref{sec:dy_prompt}.

\paragraph{Task Evaluation Metrics.}
At the task level, we adopted the following metrics: (1) Task success rate (SR): Based on the milestone success rate introduced above, a task is considered complete as long as all milestones of the task are completed; otherwise, the task is considered to have failed. (2) Average task progress (ATP): To capture task completion at a finer granularity, we quantify progress based on milestone completion and compute a model’s average progress across all tasks. (3) Milestone step ratio (MSR): To assess execution efficiency, we compare the number of steps taken by the agent to the number of human-annotated steps required to reach the same milestone.

\renewcommand{\arraystretch}{1.3} 
\begin{table*}[!t]
\renewcommand{\arraystretch}{1.1}
  \centering
  \caption{\textbf{Static Evaluation Results on \modelname.} HL denotes high-level instructions, LL denotes low-level instructions, AMS represents Action Matching Score, ATP represents Average Task Progress, and ``\# tokens" reports the average number of output tokens.}
  \small
  % \begin{tabularx}{\textwidth}{l*{13}{>{\centering\arraybackslash}X}}
  \begin{tabularx}{\textwidth}{l*{12}{>{\centering\arraybackslash}X} >{\centering\arraybackslash}p{1.1cm}}
    \toprule
    \multirow{2}{*}{\textbf{Model}} 
    & \multicolumn{2}{c}{Chinese-LL} 
    & \multicolumn{2}{c}{Chinese-HL} 
    & \multicolumn{2}{c}{English-LL} 
    & \multicolumn{2}{c}{English-HL} 
    & \multicolumn{2}{c}{Total-LL}
    & \multicolumn{3}{c}{Total-HL} \\
    \cmidrule(r){2-3}
    \cmidrule(r){4-5}
    \cmidrule(r){6-7}
    \cmidrule(r){8-9}
    \cmidrule(r){10-11}
    \cmidrule(r){12-14}
    & AMS & ATP & AMS & ATP & AMS & ATP & AMS & ATP & AMS & ATP & AMS & ATP & {\# tokens} \\
    \hline
    % \midrule
    % \rowcolor{gray!20} 
    % \multicolumn{13}{c}{\textcolor{gray}{Agent workflows}} \\
    \multicolumn{14}{c}{\textit{Agent workflows}} \\
    \hline
    Mobile-Agent-v2~\cite{wang2024mobileagentv} & 47.18 & 23.38 & 32.66 & 13.53 & 47.46 & 23.14 & 33.43 & 13.38 & 47.27 & 23.21 & 32.92 & 13.46 & - \\
    Mobile-Agent-E~\cite{wang2025mobile} & {74.52} & 38.27 & \underline{50.12} & 20.81 & \underline{75.22} & \underline{37.54} & 49.85 & 19.97 & 74.98 & 37.65 & 50.03 & 20.07 & - \\
    GPT-4o + UGround-V1-7B~\cite{gou2024navigating} & 64.15 & 31.64 & 46.08 & \underline{21.59} & 66.63 & 35.46 & 44.68 & 20.18 & 69.73 & 38.44 & 45.61 & 21.16 & - \\
    \hline
    % \rowcolor{gray!20} 
    % \multicolumn{13}{c}{\textcolor{gray}{Agent-as-a-models}} \\
    \multicolumn{14}{c}{\textit{Agent-as-a-models}} \\
    \hline
    GPT-4o~\cite{hurst2024gpt} & 29.98 & 11.90 & 23.48 & 10.45 & 28.13 & 10.56 & 23.78 & 8.27 & 29.36 & 11.50 & 23.58 & 9.79 & - \\
    Claude-3.7-sonnet~\cite{anthropic2025claude37} & 28.25 & 9.62 & 22.28 & 7.35 & 32.66 & 9.53 & 27.33 & 8.46 & 29.73 & 9.60 & 23.97 & 7.68 & - \\
    Gemini-2.5-Flash~\cite{comanici2025gemini} & 32.57 & 11.61 & 26.61 & 10.24 & 31.46 & 9.34 & 23.44 & 8.38 & 30.24 & 10.90 & 25.55 & 9.61 & - \\
    Gemini-2.5-Pro~\cite{comanici2025gemini} & 45.23 & 21.32 & 39.07 & 16.71 & 44.02 & 19.57 & 35.85 & 13.97 & 44.82 & 20.77 & 37.99 & 15.88 & - \\
    Qwen2.5-VL-7B~\cite{bai2025qwen25vl} & 64.14 & 37.64 & 34.43 & 19.91 & 50.95 & 21.47 & 26.81 & 15.63 & 59.71 & 32.73 & 31.88 & 18.62 & - \\
    \hline
    OS-Atlas-7B-Pro~\cite{OS-Atlas-Pro-7B} & 49.84 & 24.07 & 35.11 & 12.66 & 47.28 & 22.40 & 37.92 & 15.58 & 48.98 & 23.56 & 36.05 & 13.54 & - \\
    Aguvis-7B~\cite{xu2024aguvis} & 66.37 & 36.99 & 12.73 & 3.31 & 60.04 & 29.19 & 10.99 & 1.92 & 64.25 & 34.64 & 12.15 & 2.89 & \textbf{15.29} \\
    UI-Venus-Navi-7B~\cite{gu2025ui} & 64.15 & 31.64 & 45.92 & 17.62 & 55.93 & 23.58 & 45.12 & 15.27 & 61.40 & 29.21 & 45.65 & 16.91 & 44.98 \\
    UI-AGILE-7B~\cite{lian2025ui} & 59.30 & 23.66 & 38.22 & 10.92 & 55.99 & 25.73 & 38.10 & 13.74 & 58.19 & 24.29 & 38.13 & 11.77 & 50.32 \\
    AgentCPM-GUI-8B~\cite{zhang2025agentcpm} & 71.13 & 38.19 & 42.90 & 16.17 & 71.67 & 37.12 & 43.93 & 18.90 & 71.31 & 37.87 & 43.25 & 16.99 & 37.75 \\
    UI-TARS-7B-DPO~\cite{qin2025ui} &  \textbf{82.25} & \textbf{55.31} & 49.61 & 19.81 & \textbf{79.41} & \textbf{45.80} & \textbf{54.48} & \textbf{25.23} & \textbf{81.30} & \textbf{52.45} & \underline{51.24} & \underline{21.45} & \underline{22.23} \\
    UI-TARS-1.5-7B~\cite{qin2025ui} &  \underline{78.92} &  \underline{48.60} & \textbf{55.74} & \textbf{24.13} & 73.83 & 37.35 & \underline{51.27} & \underline{21.32} & \underline{77.22} & \underline{45.21} & \textbf{54.21} & \textbf{23.28} & 62.47 \\
    \bottomrule
  \end{tabularx}

  \label{tab:static_result}
  \vspace{-3mm}
\end{table*}

\renewcommand{\arraystretch}{1.3} 
\begin{table*}[!t]
\renewcommand{\arraystretch}{1.1}
  \centering
  \caption{\textbf{Dynamic Evaluation Results on \modelname.} SR denotes task success rate, AMS represents Action Matching Score, and ATP represents Average Task Progress}
  \small
  \begin{tabularx}{\textwidth}{l*{9}{>{\centering\arraybackslash}X}}
    \toprule
    \multirow{2}{*}{\textbf{Model}} 
      & \multicolumn{3}{c}{Chinese} 
      & \multicolumn{3}{c}{English} 
      & \multicolumn{3}{c}{Total} \\
    \cmidrule(r){2-4}
    \cmidrule(r){5-7}
    \cmidrule(r){8-10}
      & SR & ATP & MSR & SR & ATP & MSR & SR & ATP & MSR \\
    \hline
    % \midrule
    Mobile-Agent-v2~\cite{wang2024mobileagentv} & 6.57 & 40.49 & 1.43 & 5.88 & 41.57 & 1.23 & 6.36 & 40.83 & 1.37 \\
    Mobile-Agent-E~\cite{wang2025mobile} & \textbf{11.84} & \textbf{51.62} & \textbf{1.23} & \textbf{14.71} & \textbf{47.21} &  \underline{1.06} & \textbf{12.73} & \textbf{50.47} & \textbf{1.18} \\
    GPT-4o + UGround-V1-7B~\cite{gou2024navigating} & 6.57 & 38.21 & 1.56 &  \underline{8.82} & 44.31 & 1.27 & 7.27 & 39.65 & 1.47 \\
    \hline
    Qwen2.5-VL-7B~\cite{bai2025qwen25vl} & 0.00 & 31.81 & 1.85 & 0.00 & 28.70 & 1.71 & 0.00 & 29.89 & 1.81 \\
    UI-Venus-Navi-7B~\cite{gu2025ui} & 5.26 & 40.12 & \underline{1.32} &  \underline{8.82} & 38.77 & 1.27 & 6.36 & 39.08 & 1.31 \\
    AgentCPM-GUI-8B~\cite{zhang2025agentcpm} & 2.63 & 41.86 & 1.43 & 0.00 & 36.57 & 1.52 & 1.82 & 40.39 & 1.46 \\
    UI-TARS-7B-DPO~\cite{qin2025ui} & 5.26 & 38.27 & 1.47 & 5.88 & 40.31 & \textbf{0.95} & 5.45 & 40.06 & 1.32 \\
    UI-TARS-1.5-7B~\cite{qin2025ui} & \underline{10.53} & \underline{46.96} & 1.38 &  \underline{8.82} &  \underline{44.71} &  \underline{1.06} &  \underline{10.00} &  \underline{46.31} &  \underline{1.29} \\
    \bottomrule
  \end{tabularx}

  \label{tab:dynamic_result}
  \vspace{-3mm}
\end{table*}

\section{Experiment}

Based on the static and dynamic evaluation methods introduced in \cref{sec:eval_method}, we conducted a comprehensive evaluation of existing representative agents. We also provided a detailed analysis of several metrics within our proposed milestone dimensions, and finally offered some key insights.

\subsection{Experiment Setup.}
\paragraph{Setup.}

To reproduce the agents’ intrinsic performance as faithfully as possible, we use each model’s official prompts and tool-calling setup. In dynamic evaluation, ``GPT-4o-0806" is used as the judging model to determine milestone completion. Considering evaluation cost, we randomly selected 110 samples for dynamic evaluation, including 76 Chinese and 34 English items. To ensure stable reproducibility of the experiments, we provide the detailed configuration of each agent in Appendix \ref{sec:detail}.

\paragraph{Baseline Agents.}
We evaluated relatively stronger agents from both the Agent Workflows and Agent-as-a-Model categories, focusing specifically on those that rely solely on GUI screenshots. This restriction is necessary because many commercial apps in our benchmark do not permit access to XML layouts or other structured information due to permission constraints.
In addition, during dynamic evaluation, certain models could not be assessed because they do not support swipe operations with precise starting positions, which are required by our task design.

\subsection{Main Results}

\paragraph{Benchmark results on static evaluation.}

In \cref{tab:static_result}, we present the static evaluation results of different agents on our benchmark under both high-level and low-level instructions. The results reveal several key observations:
(1) The overall performance of all agents is quite limited. Even the best-performing model, UI-TARS-1.5-7B, achieves only a 54.21 high-level action-matching score and an average task progress of just 23.28. This indicates that agents perform poorly when dealing with environmental anomalies and fine-grained widget operations.
(2) Agent Workflows demonstrates relatively better performance. We attribute this to its additional memory module and planning module, which provide clear advantages in state awareness for long-latency tasks.
(3) Open-source GUI-fine-tuned “agent-as-a-model” systems score significantly higher under low-level instructions than under high-level ones, especially models without historical memory such as Aguvis-7B and AgentCPM-GUI. This suggests that execution-state awareness has a substantial impact on subsequent task decision-making.
(4) For open-source agents fine-tuned on GUI tasks, performance in English environments is better than in Chinese environments. We believe this is related to the training data they use and the greater interface complexity in Chinese contexts. In contrast, closed-source VLM models benefit from their extensive data and thus maintain roughly consistent performance in Chinese environments.

\paragraph{Benchmark results on dynamic evaluation.}
Table \cref{tab:dynamic_result} shows the execution results of different agents on real devices, with a maximum of 50 steps per attempt. The results indicate that the best-performing agent in task execution is Mobile-Agent-E, but its task success rate is only 12.73. The findings show that:
(1) Similar to the static evaluation, we observe that Agent Workflows—equipped with full memory, planning, and reflection—performs significantly better than agent-as-a-model systems.
(2) We find that during dynamic execution, compared to the static evaluation, the models’ ATP scores improve substantially. We believe this is because in dynamic evaluation, models can retry and correct mistakes multiple times. Notably, UI-TARS-1.5-7B even performs worse than UI-TARS-7B-DPO in some static reasoning results, but the former performs significantly better in dynamic reasoning.
(3) In Chinese environments, the milestone-step ratio is significantly higher than in English environments, which reflects the greater complexity of UI layouts and navigation logic in Chinese apps.
(4) Although all models exhibit low task success rates (SR), the average task progress provides a more informative distinction between models’ performance, and its trend aligns with that of SR.

\subsection{Analysis}

\begin{figure}[t]
    \centering
    \includegraphics[width=0.95\linewidth]{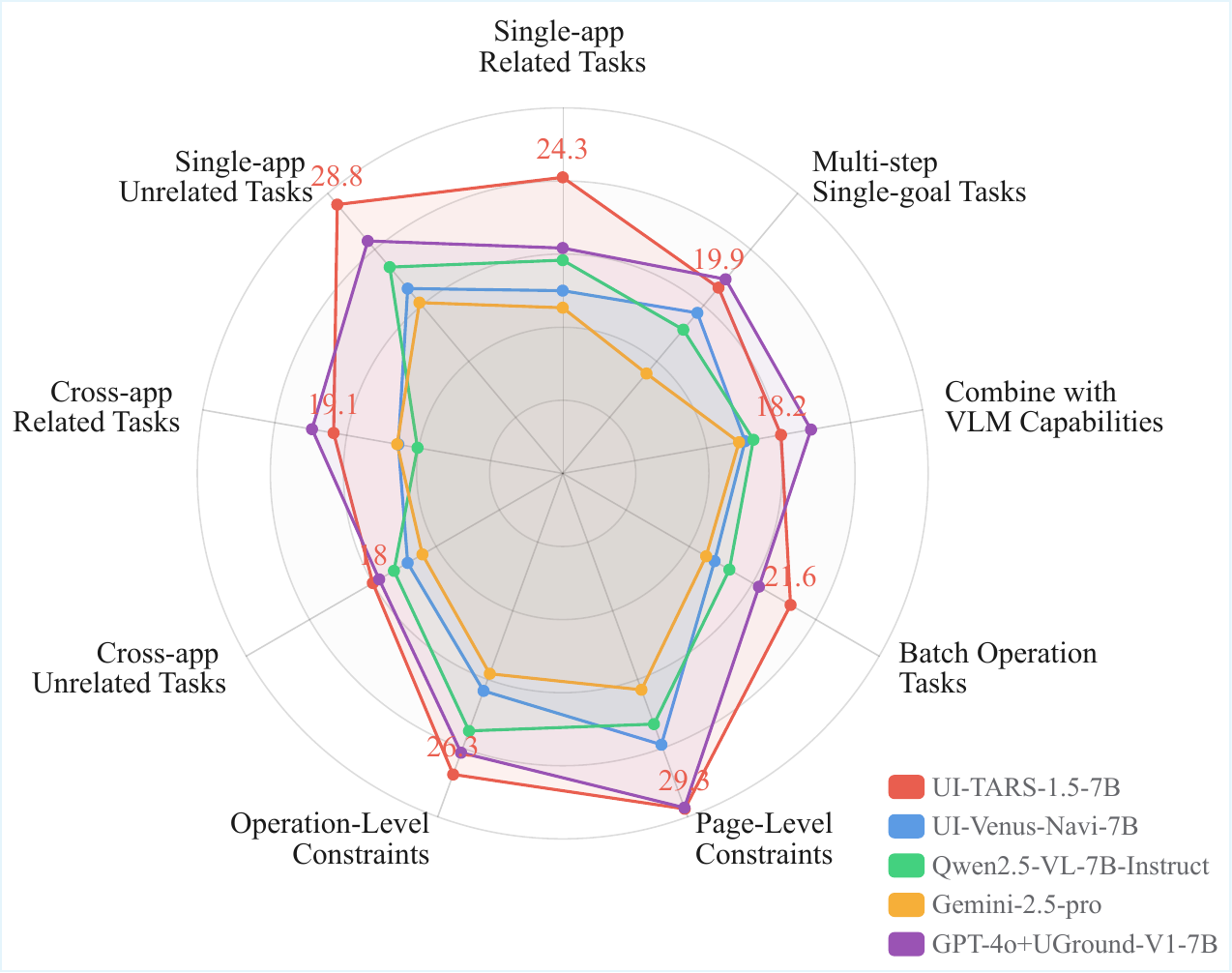}
    % \vspace{-0.4cm}
    \caption{\textbf{Action matching scores of high-level instructions across different task categories.} The range of all axes is normalized to 0-30.}
    \label{fig:task_types}
    % \vspace{-0.4cm}
\end{figure}

\begin{figure}[t]
    \centering
    \includegraphics[width=0.86\linewidth]{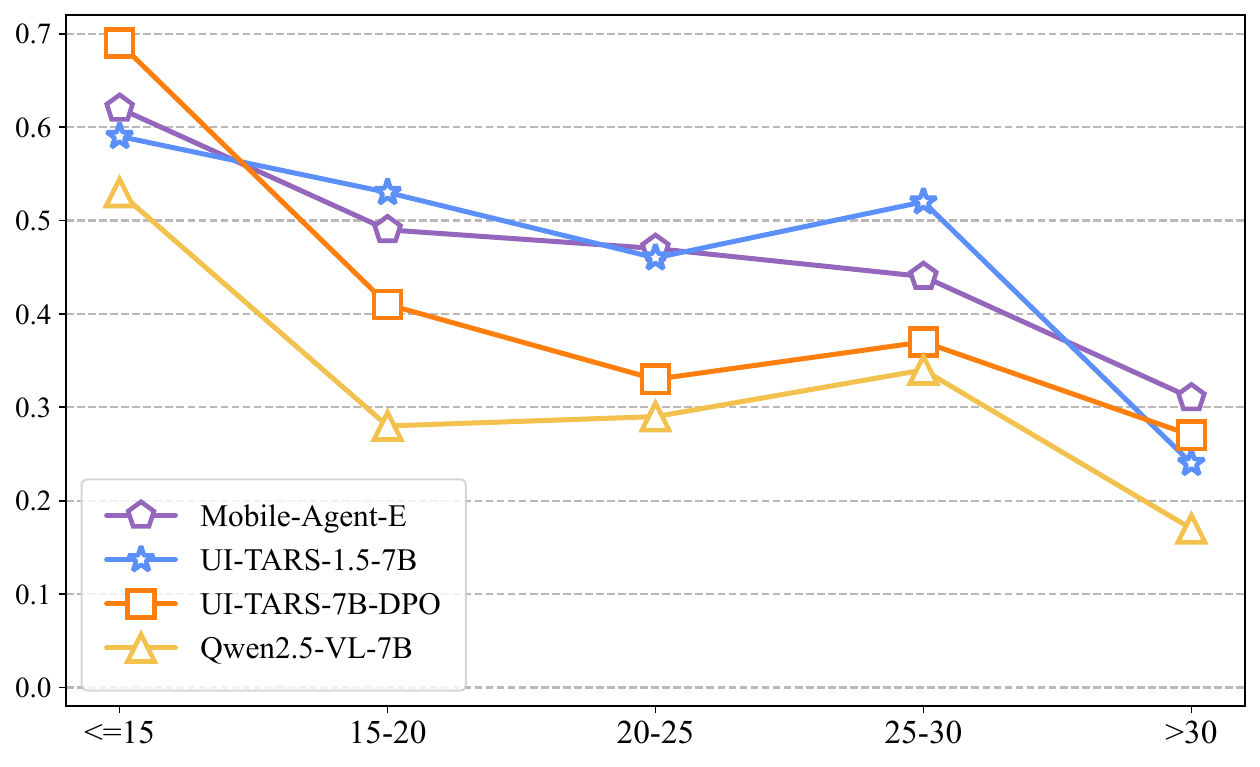}
    % \vspace{-0.4cm}
    \caption{\textbf{Average task progress curves of agents under different step lengths.}}
    \label{fig:diff_line}
    % \vspace{-0.4cm}
\end{figure}

\begin{figure}[!h]
    \centering
    \includegraphics[width=\linewidth]{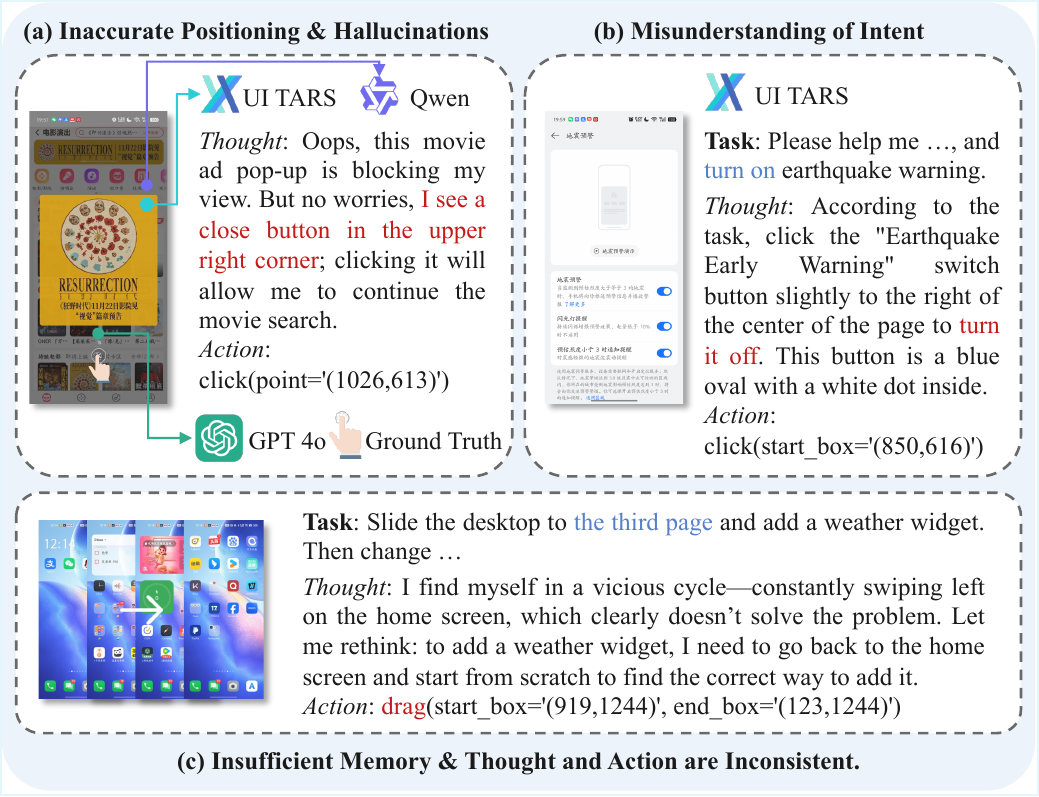}
    % \vspace{-0.4cm}
    \caption{\textbf{Qualitative analysis of several typical errors.}}
    \label{fig:case}
    % \vspace{-0.4cm}
\end{figure}

\paragraph{Refined components pose challenges to agent perception.}
Existing models, especially GUI-specialized models, already have strong abilities in understanding and locating on-screen information and have achieved notable results on current benchmarks~\cite{li2024effects,zhang2024android,lu2025guiodyssey}. However, as shown in \cref{tab:static_result}, their action-matching performance on our benchmark is not ideal. In Appendix \ref{sec:detail}, we additionally report the type-matching and action-matching scores for different actions. For the relatively strong UI-TARS-1.5-7B model, the CLICK action type-matching score reaches 94.45, but the action-matching score is only 62.34. This indicates that when faced with the richer and more diverse UI widgets in real environments, the model still struggles with precise localization, such as the lightly colored close button shown in \cref{fig:case} (a).
We also believe that swipe operations require further exploration. As shown in \cref{fig:overview}, various types of sliding components are generally difficult for current models to manipulate with fine control. Specific examples are provided in Appendix \ref{sec:case}.

\paragraph{Performance analysis under different task types.}

We evaluate model performance according to the task categories described in \cref{sec:data_stat}. As shown in \cref{fig:task_types}, we find that the models achieve lower scores on multi-goal tasks. Our analysis shows that, in particular, for cross-application tasks, the models struggle to maintain a clear memory of past states when switching between different apps. They may even misunderstand the task itself, as illustrated in \cref{fig:case} (b).
We also observe that tasks requiring the integration of large-model capabilities have similarly low success rates. The analysis indicates that agents often become confused about whether they should rely on phone operations or their own reasoning abilities. For example, in translation tasks, although the model itself has the ability to translate, it tends to open the phone’s translation app instead, as shown in Appendix \ref{sec:case}.

\paragraph{Performance analysis under different task difficulties.}
To evaluate how task execution length affects model performance, we divide tasks into five categories based on the number of steps required for humans to complete them, using intervals of five steps. As shown in \cref{fig:diff_line}, we report the dynamic evaluation results of several representative models under different difficulty groups. Across these groups, the models’ average task progress decreases as the number of required steps increases. This indicates that models typically complete only the initial milestones of a task, and that long-latency context loss and error accumulation make it difficult to complete later stages. Insufficient memory modules can even cause models to fall into infinite loops, as illustrated in \cref{fig:case} (c).

\paragraph{The impact of thinking patterns on agent performance.}
The influence of reasoning modes on agent performance. Chain-of-thought reasoning has been widely applied across many domains, and current GUI agents also commonly rely on this approach to enhance their reasoning capabilities. However, we find that when agents encounter pages that do not match their expectations, they often become confused or even fall into infinite loops. Their ability to reflect, explore, and guide themselves back onto the correct trajectory remains insufficient. Inconsistencies between reasoning and actions further limit the performance of GUI agents, as shown in \cref{fig:case} (c). Moreover, as indicated in Table \cref{tab:static_result}, the reasoning mode introduces additional token costs, and improving the efficiency of agents on mobile devices is also an important concern.

In addition, we conducted several supplementary explorations to further reveal the model’s capabilities and limitations. As shown in Appendix~\ref{sec:detail}, the long-tail distribution of action types significantly impacts task success. Appendix~\ref{sec:ad_case} illustrates the model’s ability to handle real-world interruptions such as ads, authorization prompts, and login pop-ups. Moreover, many commercial apps require rapid secondary confirmations (e.g., exiting an app), which current agents often fail to complete due to inference latency and limited action spaces, as shown in Appendix~\ref{sec:case}.
\section{Conclusion}

We introduce \modelname, a comprehensive framework for evaluating long-latency smartphone tasks, covering 571 complex tasks across 38 domains in both Chinese and English. We provide high-quality human trajectories and mitigate static-evaluation bias through multi-trajectory verification. For dynamic evaluation, where current agents still struggle with complex tasks, we propose a milestone-based dynamic metric, Average Task Progress (ATP), to enable fine-grained performance analysis. Experimental results show that even the most powerful agent achieves only an ATP of 50.47 on \modelname.
Our analysis highlights four key areas for improvement: richer awareness of small on-screen widgets, finer-grained action control, more effective long-term memory, and stronger error-recovery abilities.

{
    \small
    \bibliographystyle{ieeenat_fullname}
    \bibliography{main}
}

\clearpage
\setcounter{page}{1}
\maketitlesupplementary

\appendix

The appendix provides additional details and supplementary information to further elaborate on the sections above. \cref{sec:action} presents the action space defined by our benchmark. \cref{sec:all_app} includes all the everyday scenarios involved as well as the selected representative apps. \cref{sec:cross_app} provides a taxonomy of cross-app tasks along with examples. \cref{sec:statis} analyzes the statistical characteristics of the dataset. \cref{sec:dy_prompt} describes the prompt used for milestone determination during dynamic evaluation. \cref{sec:detail} details the evaluation experiments, including benchmark evaluation specifics and the Agent’s evaluation settings. \cref{sec:case} provides examples of tasks from different categories and also showcases some sample model outputs. \cref{sec:ad_case} specifically provides examples of various agent responses to real-world emergencies.

\section{Action Set}\label{sec:action}
Our trajectory annotation primarily uses the Android Debug Bridge (ADB) to connect to physical devices for GUI navigation. During this process, we capture device states and detailed information about operations, such as the coordinates of click events and the triggering of function keys. The details of our action set are shown in \cref{tab:action_space}. The ``Swipe" action specifies explicit start and end points.

\begin{table*}[ht]
\centering
\caption{\textbf{Definition of Action Space.}}
\label{tab:action_space}
\renewcommand{\arraystretch}{1.2}
\resizebox{0.99\textwidth}{!}{
\begin{tabular}{l l}
\toprule
\textbf{Action} & \textbf{Definition}          \\
\midrule
Click(x, y)  & Click the point on the screen with coordinate (x, y).      \\
Swipe(x1, y1, x2, y2) & Swipe from the starting point with coordinate (x1, y1) to the end point with coordinates2 (x2, y2).      \\
LongPress(x, y)  &  Press the point on the screen with coordinate (x, y) for specified seconds.     \\
Type(text)  &  Input the specified text into the activated input box.     \\
PressBack()  &  Press the system button $Back$.    \\
PressHome()  &  Press the system button $Home$.    \\
PressMenu()  &  Press the system button $Menu$.    \\
Terminate('success' or 'failure')  &  Terminate the current task and report its completion status, e.g., $success, failure$. \\
Wait()  &  Wait specified seconds for the change to happen. \\
\bottomrule
\end{tabular}}
\end{table*}

\section{All Scenarios and Apps}\label{sec:all_app}

To best reflect the agent’s effectiveness on long-horizon tasks in everyday life, we identified 38 scenarios through discussion, encompassing 62 commercial apps and 12 system apps in both Chinese and English settings. Details are shown in \cref{tab:all_app}, which also reports the number of long-horizon tasks associated with each app.

\begin{table*}[t!]
\centering
\renewcommand{\arraystretch}{1.25} 
\caption{\textbf{All Scenarios and Apps.}}
\begin{tabular}{lcccc}
    \toprule
    \textbf{Domain} & \textbf{Chinese APP} & \textbf{\# Task}   & \textbf{English APP} & \textbf{\# Task} \\
    \hline
    % \midrule
    \rowcolor{gray!15}
    \multicolumn{5}{l}{\emph{Commercial Applications}}    \\ 
    Online Shopping    & Taobao, JD.com  &  56 & eBay, AliExpress  & 23   \\
    Mobile Payments    & Alipay          &  18 & PayPal  &  2  \\
    Instant Messaging  & WeChat, QQ      &  95  & Facebook, Telegram  & 22    \\
    Lifestyle Sharing  & Xiaohongshu     &  24 & Instagram  &  13  \\
    Knowledge Sharing  & Zhihu           &  20 & Quora  & 5   \\
    Social Media       & Weibo           &  16 & X  & 4   \\
    Video Sharing      & Bilibili        &  21 & YouTube  & 14   \\
    Short Video        & Douyin          &  26 & TikTok  & 2   \\
    Local Services     & Meituan, Fliggy &  55 & Trip.com, Booking.com  &  21  \\
    Navigation \& Transportation & Amap, DiDi &  32 & Google Maps, Uber  &  12  \\
    Search             & Baidu           &  27 & Google Chrome  & 13   \\
    Digital Reading    & Tomato Novel    &  16 & WebNovel  &  5  \\
    Online Meetings    & Tencent Meeting &  10 & Zoom  & 4   \\
    News Aggregation   & Toutiao         &  9 & Google News  & 5   \\
    Cloud Storage      & Alibaba Cloud Drive &  9 & Google Drive  & 10   \\
    Office Tools       & WPS Office      &  16 & Microsoft Office  & 2   \\
    Logistics Tracking & Cainiao         &  3 & 17Track  & 1   \\
    Secondhand Marketplace & Xianyu      &  12 & eBay  & 8   \\
    Workplace Collaboration & Feishu, DingDing &  18 & Teams  & 7   \\
    Music Streaming    & QQ Music        &  16 & Spotify  & 8   \\
    Video Streaming    & Youku           &  24 & Prime Video  & 4   \\
    Task Management    & DIDA            &  12 & Microsoft To Do  & -   \\
    Local Reviews      & Dianping        &  10 & Yelp  & 5   \\
    Job Recruitment    & BOSS Zhipin     &  1 & LinkedIn  &  4  \\
    Fitness \& Health  & KEEP            &  9 & MyFitnessPal  & 2   \\
    App Store          & App Store       &  4 & Play Store  & -   \\
    \hline
    % \midrule
    \rowcolor{gray!15}
    \multicolumn{5}{l}{\emph{System Application}}    \\ 
    Weather            & Weather         &  17 & Weather  & 7   \\
    Calculator         & Calculator      &  13 & Calculator  & 10   \\
    Calendar           & Calendar        &  21 & Calendar  &  9  \\
    Clock              & Clock           &  5 & Clock  &  5  \\
    Messages           & Messages        &  9 & Messages  & 2   \\
    Contacts           & Contacts        &  10 & Contacts  & 8   \\
    Photos             & Photos          &  15 & Photos  &  15 \\
    Mail               & Mail            &  11 & Mail  &  6  \\
    Notes              & Notes           &  19 & Notes  & 11   \\
    Recorder           & Recorder        &  2 & Recorder  &  1  \\
    My Files           & My Files        &  9 & My Files  & 18   \\
    Settings           & Settings        &  4 & Settings  &  12  \\
    \bottomrule
\end{tabular}
\label{tab:all_app}
\end{table*}

\section{Cross APP Tasks}\label{sec:cross_app}
As the functionalities of various apps grow increasingly complex, cross-app interaction patterns have also become more diverse. To more comprehensively evaluate an agent’s performance in scenarios involving frequent switches between different app interfaces, we incorporate a variety of typical cross-app interaction types into our benchmark, including sharing, clipboard operations, application redirection, authorization, tool call, comparison, information aggregation, and multi-round cross-app interactions. For each interaction type, we provide corresponding examples in the benchmark, as shown in \cref{tab:cross_app}.

\begin{itemize}
\item Sharing:  
After completing an operation in one app, pass the result to another app via the app’s built‑in sharing feature or by sharing it manually.
\item Clipboard Operation:  
Transfer text or other content between different apps using copy, paste, and similar actions.
\item Application Redirection:  
When obtaining information or tapping an entry point in one app, the system automatically redirects and opens another app.
\item Authorization:  
Use an existing account in one app to complete third‑party authorization login or account binding in another app.
\item Tool Call:  
(1) Modify the features or status of other apps through system settings.  
(2) Invoke built‑in phone tools or system apps (such as camera, contacts, etc.) within an app to complete related operations.
\item Comparison:  
Collect similar types of information from multiple apps, compare them, and then make a choice and perform follow‑up actions based on the comparison results.
\item Information Aggregation:  
Collect and organize information on one or more platforms, then consolidate it into another app for recording, archiving, or unified management.
\item Multi-Round Interactions:  
Need to switch back and forth between multiple apps multiple times, completing the task step by step, where subsequent steps depend on the results of previous steps.
\end{itemize}

\begin{table*}[ht]
\centering
\caption{\textbf{Classification and examples of cross-application interaction tasks.}}
\label{tab:cross_app}
\renewcommand{\arraystretch}{1.2} 
% \resizebox{0.99\textwidth}{!}
{
\begin{tabular}{l p{0.74\textwidth}}
\toprule
\textbf{Classification} & \textbf{Example}          \\
\midrule
Sharing & Help me find a nearby cinema on Meituan and buy two tickets for a popular movie, choosing two adjacent seats as close to the middle as possible. After purchasing, \underline{take a screenshot of the ticket information and send it to ``GUI Test" on QQ}. \\
Clipboard Operation & Open Google Chrome and search for AI answers to ``A healthy lunch suitable for the beginning of autumn". Choose one recommended dish, \underline{copy the ingredient list from its recipe into my Notes app}, then search YouTube for a video on how to make it. \\
Application Redirection & Open Trip, search for food, set the location to International Trade CBD, apply the Open now filter, open the nearest place, \underline{click its map location, then switch to Google Maps} and show driving directions from International Trade CBD to that place. \\
Authorization & Go to the app store to install Zhihu. After the installation, open the app and complete the login process \underline{by authorizing via Weibo}. \\
Tool Call & Open my phone's Settings and go to Emergency SOS; turn on quick emergency access and automatic location sharing, \underline{add Dad and Mom from my contacts} as emergency contacts, and then enable Earthquake Early Warning. \\
Comparison & On Yelp, search for coffee shops near the Googleplex, filter to Open Now and within 1 mile, and open the top two listings. Then in Google Maps, get walking directions from the Googleplex to each, \underline{pick the one with the shortest walk time}, and save that business on Yelp. \\
Information Aggregation & Go to Quora, search for the query ``Chinese food recommend", review at least five posts, \underline{identify the three dishes mentioned most often}, and add them as separate lines in a new note in the Notes app titled ``Chinese food summary". \\
Multi-Round Interactions & Open Uber Eats, set my delivery address to the Googleplex, then search for Pizza, Poke, Fast Food, and Chinese Food. \underline{For each category}, pick one restaurant with a delivery ETA under 30 minutes and \underline{send the links to my Telegram contact ``liuyujue"}. After you've shared all four, send message: ``Which restaurant shall we go to tonight?" \\

\bottomrule
\end{tabular}}
\end{table*}

\section{Data Statistics}\label{sec:statis}

\begin{figure}[t]
    \centering
    \includegraphics[width=0.86\linewidth]{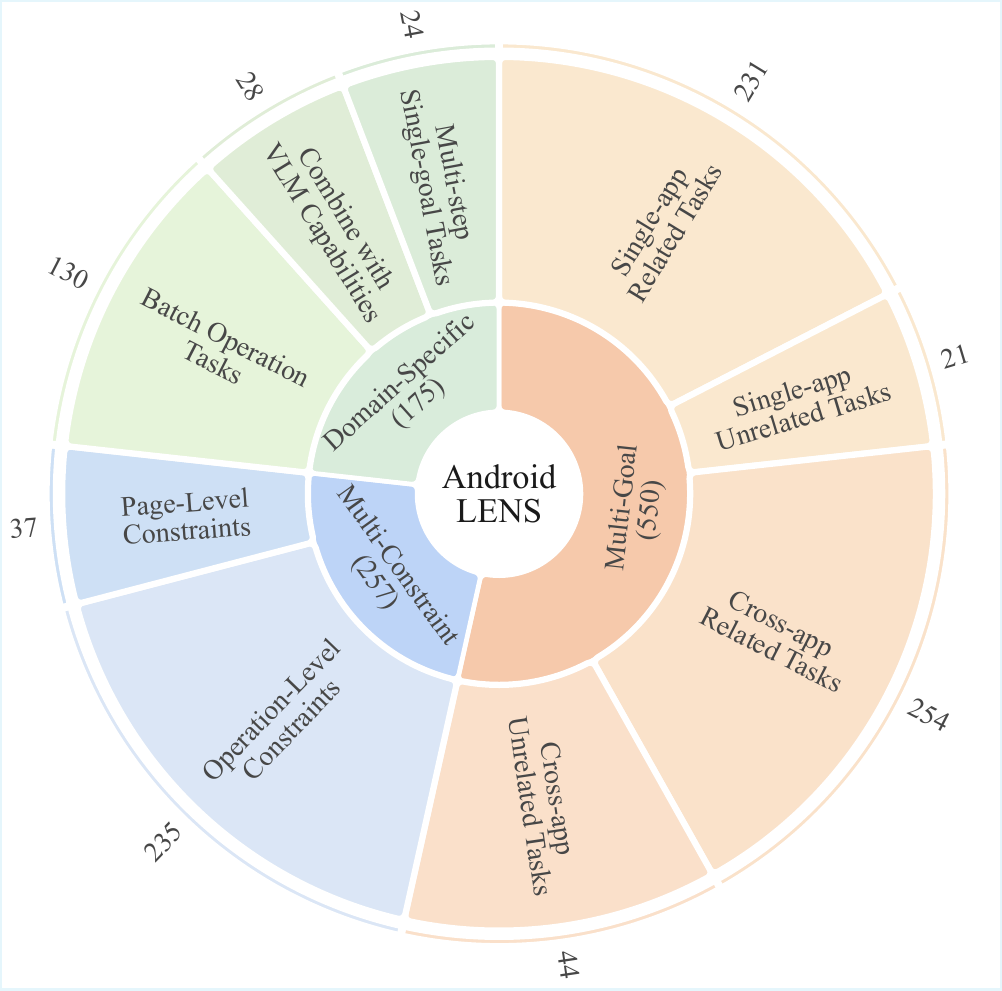}
    % \vspace{-0.4cm}
    \caption{\textbf{Distribution of different categories in \modelname.}}
    \label{fig:bench_dist}
    % \vspace{-0.4cm}
\end{figure}

\begin{figure}[t]
    \centering
    \includegraphics[width=0.95\linewidth]{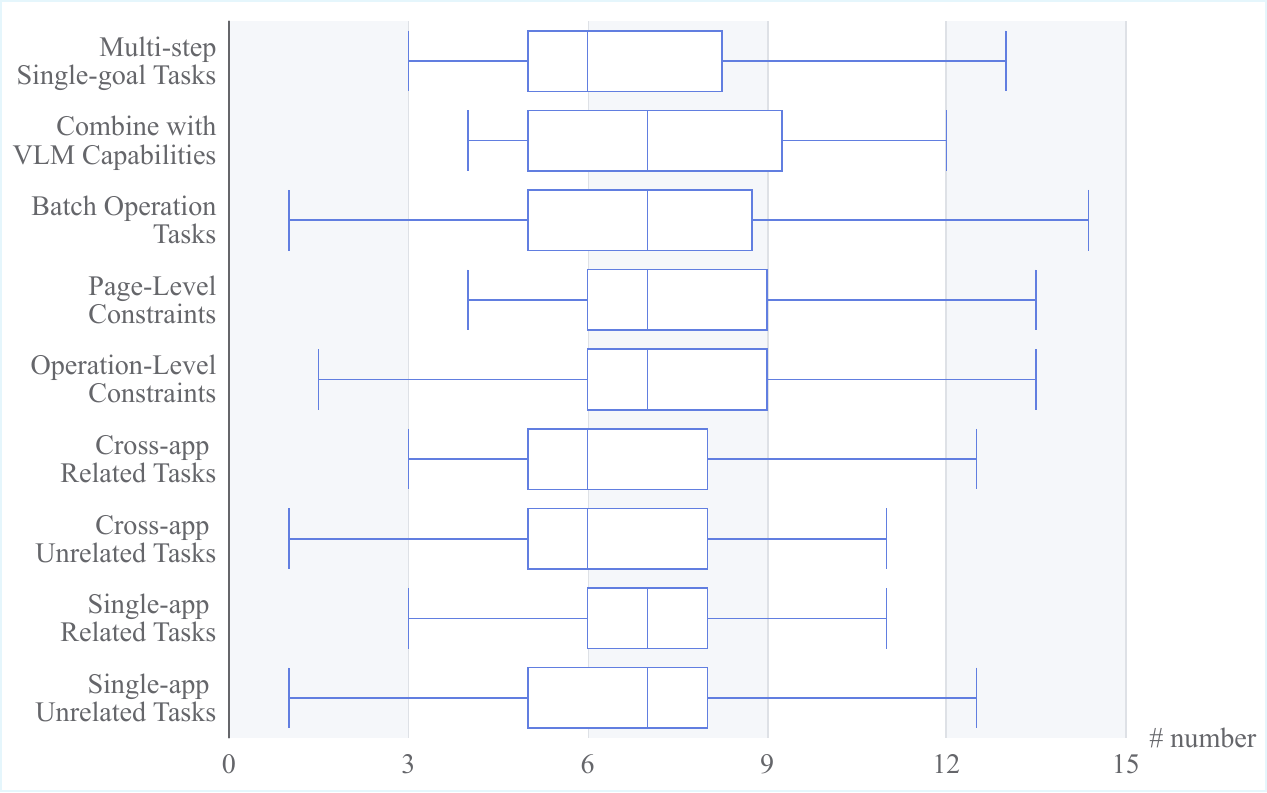}
    % \vspace{-0.4cm}
    \caption{\textbf{Distribution of milestone numbers across different categories.}}
    \label{fig:mile_dist}
    % \vspace{-0.4cm}
\end{figure}

Overall, \modelname comprises 399 Chinese tasks and 172 English tasks, including 273 single-application tasks and 298 cross-application tasks. As described in \cref{sec:data_stat}, we divide all tasks into three major categories: Multi-constraint, Multi-goal, and Domain-specific. For each major category, we further define several subcategories, and the distribution of the number of tasks in each subcategory is shown in \cref{fig:bench_dist}. Meanwhile, we assign a different number of milestones to each task to characterize its execution difficulty, and their distribution within each subcategory is shown in \cref{fig:mile_dist}.

\section{Evaluation Prompts}\label{sec:dy_prompt}

\subsection{General Model Evaluation Prompt}

To showcase each model’s capabilities on GUI tasks as fully as possible, we directly used the official configuration to guide models that provide an official GUI navigation system prompt. For general-purpose models that do not offer a dedicated GUI prompt, such as GPT‑4o and Claude 3.7 Sonnet, we referred to the prompt design of UI-TARS-1.5-7B and constructed a unified, similar system prompt. The specific content is as follows.

\begin{bluepromptbox}{System Prompt}
You are a GUI agent. You are given a task and your action history, with screenshots. You need to perform the next action to complete the task. \\ \\
\#\# Output Format \\

Thought: ... \\
Action: ... \\

\#\# Action Space \\
click(start\_box=`\textless\textbar box\_start\textbar\textgreater(x1,y1)\textless\textbar box\_end\textbar\textgreater') \\
long\_press(start\_box=`\textless\textbar box\_start\textbar\textgreater(x1,y1)\textless\textbar box\_\\end\textbar\textgreater', time=`') \\
type(content=`') \\
scroll(start\_box=`\textless\textbar box\_start\textbar\textgreater(x1,y1)\textless\textbar box\_end\textbar\textgreater', direction=`down or up or right or left') \\
press\_back() \\
press\_home() \\
wait() \\
finished() \# Submit the task regardless of whether it succeeds or fails. \\ \\
\#\# Note \\
- Use \{\textit{language}\} in Thought part. \\
- Summarize your next action (with its target element) in one sentence in Thought part. \\\\
\#\# User Instruction \\
\{\textit{instruction}\}

\end{bluepromptbox}

The placeholders \{\textit{language}\} and \{\textit{instruction}\} will be dynamically replaced according to the specific properties of each task. For actions that involve coordinate information (x1, y1), we adopt the coordinate format preferred by each general model: GPT-4o and Claude-3.7-Sonnet use absolute coordinates, while Gemini-2.5-Flash and Gemini-2.5-Pro use relative coordinates in the range 0–1000.

Since our benchmark also evaluates low-level instructions, in such cases we explicitly append the low-level instruction to the prompt in the following format:
``
\#\# The action you need to take now\textbackslash n
\{\textit{low\_instruction}\}
"
This ensures that the model can accurately reference the corresponding low-level instruction when performing the current action.

\subsection{Milestone Evaluation Pipeline}

During dynamic evaluation, we measure task progress by determining whether each milestone has been achieved, allowing us to assess task completion at a finer granularity. Specifically, we use screenshots of key components associated with each milestone, textual descriptions, and UI states as criteria to segment the agent’s reasoning and action screenshot trajectory, and then determine milestone completion segment by segment.

By default, we segment trajectories in chronological order (e.g., each segment contains 10 actions). This is because, even if some milestones are unordered, such unordered relationships are usually limited to a small subset of milestones, while the overall process is still constrained within a certain time interval.

For ordered milestones, after providing a trajectory segment and its screenshot, we feed only the information for this single milestone into GPT-4o, asking it to determine whether the subtask corresponding to the current milestone has been completed:
(1) If it is deemed completed, GPT-4o returns the index of the “last completed step.” We then start from the step after that index and select the next trajectory segment to evaluate the next milestone.  
(2) If it is deemed not completed, we start from the last step of the current trajectory segment, select the next trajectory segment, and continue evaluating the current milestone.

For unordered milestones, we input all milestones in the unordered group into GPT-4o at once. GPT-4o then determines which milestones in the group have been completed within the current trajectory segment and returns the index of the last completed step for each completed milestone. We then choose the latest (largest) index among them as the starting point for the next segmentation.
The prompt used in this process can be referred to in the following text:

\begin{greenpromptbox}{Milestone Evaluation Prompt}

You are an evaluation expert for smartphone operation tasks. Based on the provided sequence of screenshots showing the task being performed on a phone, determine whether each milestone in the specified milestone list has been successfully completed. Note that there is no required execution order among the milestones. If a milestone has been successfully completed, you need to return the index (starting from 0) of the last step in the trajectory at which that milestone is completed.\\

\#\# Guidelines \\

1. Do not make any assumptions: Judge only based on the provided screenshots. Do not infer or assume any information that is not explicitly shown in the screenshots. \\
2. Milestone completion criteria: A milestone is considered successfully completed only when:\\
   - There is no correlation between multiple milestones; determining the completion of one milestone does not require consideration of other milestones.
   - The key component corresponding to the milestone (page, button, icon, text, etc.) clearly appears in the task trajectory.\\
   - The state of that key component exactly matches the expected state required by the milestone.\\
3. Common failure reasons:\\
   - Not completed: If, throughout the entire operation trajectory, the key component or required state specified by the milestone never appears, then the milestone is considered not executed or not completed.\\
   - Noun/entity mismatch: If the milestone requires “open Taobao,” but the screenshots show the “Meituan” page, the milestone fails. Entities that look similar but are not exactly the same (for example, “Xiaomi 15” vs “Xiaomi 16,” or “driving route” vs “walking route”) are regarded as different entities and should be judged as failures.\\
   - Verb/action mismatch: If the milestone requires “like the post,” but the screenshots show “favorite the post,” the operation does not match and should be judged as a failure.\\
   - State mismatch: If the milestone requires “follow this user,” but the screenshot shows that the user is already in the “Following” state, and then the user performs an action that changes it to “Not following,” the final state does not match the requirement and should be judged as a failure.\\
4. Corrective actions: If an incorrect operation occurs at some intermediate step but is later corrected by subsequent actions, as long as the final state exactly matches the correct state required by the milestone, it is considered successfully completed.\\

\#\# Input format \\

The milestone list in the input is structured as follows:\\

[\\
  \{\\
    ``idx": 0,\\
    ``sub-target": ``Open Google Chrome and search for `panda'",\\
    ``screenshot": <image>,  \\
    ``bbox": [0.023, 0.121, 0.976, 0.189],\\
    ``text": ``panda",  \\
    ``state": [``selected"]  \\
  \},\\
  ...\\
]\\

- ``idx": A unique identifier for each milestone (starting from 0).\\
- ``sub-target": Describes the target of the current milestone.\\
- ``screenshot": A reference screenshot of the final result for this milestone.\\
- ``bbox": The location of the key component in the screenshot, represented by normalized coordinates [x1, y1, x2, y2].\\
- ``text": The text content of the key component recognized by OCR.\\
- ``state": The state information of the key component related to the milestone.\\

\#\# Output format \\

Please return a JSON list, where each item indicates whether the input milestone has been completed, similar to:\\

[\\
\{\\
~~~~``idx": 0,\\
~~~~``state": 0/1,\\
~~~~``last\_idx": 3\\
\}\\
...\\
]\\

- ``idx": The sequence number of the corresponding milestone (provided in the input).\\
- ``state" indicates the completion status of the milestone: 1 for success, 0 for failure.\\
- ``last\_idx" indicates the index of the last step in completing the milestone (starting from 0). If the milestone is not complete, -1 is returned.\\

Please strictly adhere to the JSON format for returning results, without any additional content or explanation.\\

\#\# Notes\\

- Do not infer or supplement information that is not shown in the screenshots; only judge based on visible content.\\
- The screenshots provided for each milestone are for reference only. A milestone should not be considered incomplete just because the exact same screen does not appear in the execution trace, since the phone UI may vary due to version differences, settings, or device state. \\
- You must ensure that every entity and every action described in the milestone is accurately matched, and that the final state is completed.\\
- Focus on the final state required by the milestone; do not be distracted by irrelevant or minor details in intermediate steps.\\
- Pay close attention to subtle interface differences, such as whether a tab is highlighted, or whether the text or icon of a button has changed, as these may be key to determining success or failure.\\

\end{greenpromptbox}

Through this process, we can determine the completion status of each milestone one by one. For a given task, once all milestones are judged to be completed, we consider the task as finished. Compared with directly feeding an entire, very long action trajectory into GPT-4o in a single pass, this segmented dynamic evaluation strategy effectively reduces errors in long-sequence understanding by the judgment model, thereby improving the reliability and accuracy of the evaluation.

\section{Experimental Details}\label{sec:detail}
To better analyze the causes of the models’ performance limitations, we additionally report the results for Type Matching (TM) in \cref{tab:tm_em}. The results show that many models struggle to accurately grasp the current stage of the task, leading to prediction errors. However, once they are explicitly informed of “what needs to be done in the current step,” their performance improves significantly. This indicates that existing models still have substantial shortcomings in utilizing historical context and in analyzing and reasoning about tasks.

\renewcommand{\arraystretch}{1.3} 
\begin{table*}[!t]
\renewcommand{\arraystretch}{1.1}
  \centering
  \caption{\textbf{Static Evaluation Results on \modelname.} HL denotes high-level instructions, LL denotes low-level instructions, AMS represents Action Matching Score, and TM represents action type matching.}
  \small
  % \begin{tabularx}{\textwidth}{l*{13}{>{\centering\arraybackslash}X}}
  \begin{tabularx}{\textwidth}{l*{12}{>{\centering\arraybackslash}X}}
    \toprule
    \multirow{2}{*}{\textbf{Model}} 
    & \multicolumn{2}{c}{Chinese-LL} 
    & \multicolumn{2}{c}{Chinese-HL} 
    & \multicolumn{2}{c}{English-LL} 
    & \multicolumn{2}{c}{English-HL} 
    & \multicolumn{2}{c}{Total-LL}
    & \multicolumn{2}{c}{Total-HL} \\
    \cmidrule(r){2-3}
    \cmidrule(r){4-5}
    \cmidrule(r){6-7}
    \cmidrule(r){8-9}
    \cmidrule(r){10-11}
    \cmidrule(r){12-13} 
    & AMS & TM & AMS & TM & AMS & TM & AMS & TM & AMS & TM & AMS & TM  \\
    \hline
    % \midrule
    % \rowcolor{gray!20} 
    % \multicolumn{13}{c}{\textcolor{gray}{Agent workflows}} \\
    \multicolumn{13}{c}{\textit{Agent workflows}} \\
    \hline
    Mobile-Agent-v2          & 47.18 & 63.40 & 32.66 & 59.76 & 47.46 & 64.35 & 33.43 & 55.88 & 47.27 & 63.72 & 32.92 & 58.46 \\
Mobile-Agent-E           & 74.52 & 95.36 & 50.12 & 81.26 & 75.22 & 95.31 & 49.85 & 78.12 & 74.98 & 95.34 & 50.03 & 80.21 \\
GPT-4o+UGround-V1-7B     & 64.15 & 94.95 & 46.08 & 73.23 & 66.63 & 91.68 & 44.68 & 71.87 & 69.73 & 93.43 & 45.61 & 72.77 \\
    \hline
    % \rowcolor{gray!20} 
    % \multicolumn{13}{c}{\textcolor{gray}{Agent-as-a-models}} \\
    \multicolumn{13}{c}{\textit{Agent-as-a-models}} \\
    \hline
    GPT4o                    & 29.98 & 79.08 & 23.48 & 80.53 & 28.13 & 78.18 & 23.78 & 73.03 & 29.36 & 78.78 & 23.58 & 78.02 \\
Claude-3.7-sonnet        & 28.25 & 83.52 & 22.28 & 77.76 & 32.66 & 83.58 & 27.33 & 78.20 & 29.73 & 83.54 & 23.97 & 77.91 \\
Gemini-2.5-Flash         & 32.57 & 68.83 & 26.61 & 60.59 & 31.46 & 67.27 & 23.44 & 66.49 & 30.24 & 68.31 & 25.55 & 62.57 \\
Gemini-2.5-Pro           & 45.23 & 87.95 & 39.07 & 76.55 & 44.02 & 86.53 & 35.85 & 72.01 & 44.82 & 87.48 & 37.99 & 75.03 \\
Qwen2.5-VL-7B            & 64.14 & 85.89 & 34.43 & 53.90 & 50.95 & 71.22 & 26.81 & 40.34 & 59.71 & 80.97 & 31.88 & 49.36 \\
    \hline
    OS-Atlas-7B-Pro          & 49.84 & 79.37 & 35.11 & 71.80 & 47.28 & 74.91 & 37.92 & 71.04 & 48.98 & 77.88 & 36.05 & 71.55 \\
Aguvis-7B                & 66.37 & 97.66 & 12.73 & 36.10 & 60.04 & 93.84 & 10.99 & 24.42 & 64.25 & 96.38 & 12.15 & 32.19 \\
UI-Venus-Navi-7B         & 64.15 & 83.67 & 45.92 & 75.76 & 55.93 & 79.07 & 45.12 & 75.58 & 61.40 & 82.13 & 45.65 & 75.70 \\
UI-AGILE-7B              & 59.30 & 80.11 & 38.22 & 76.46 & 55.99 & 81.51 & 38.10 & 73.17 & 58.19 & 80.58 & 38.13 & 75.36 \\
AgentCPM-GUI-8B          & 71.13 & 94.49 & 42.90 & 78.29 & 71.67 & 95.47 & 43.93 & 75.28 & 71.31 & 94.82 & 43.25 & 77.28 \\
UI-TARS-7B-DPO           & 82.25 & 97.72 & 49.61 & 77.60 & 79.41 & 95.79 & 54.48 & 76.74 & 81.30 & 97.07 & 51.24 & 77.31 \\
UI-TARS-1.5-7B           & 78.92 & 98.50 & 55.74 & 83.34 & 73.83 & 97.81 & 51.27 & 79.97 & 77.22 & 98.27 & 54.21 & 82.21 \\
    \bottomrule
  \end{tabularx}

  \label{tab:tm_em}
  \vspace{-3mm}
\end{table*}

\section{Trajectory Examples}\label{sec:case}
As shown in \cref{tab:exmp_class}, we further divide the three major categories under \modelname and provide corresponding example tasks for reference. It is worth noting that, for multi-step single-goal tasks, we focus on tasks that pursue a single objective within a single application, but require many steps due to the inherent complexity of the goal. Such tasks typically involve multiple page navigations, form filling, and similar operations.

\begin{table*}[ht]
\centering
\caption{\textbf{Classification and examples of tasks in \modelname.}}
\label{tab:exmp_class}
\renewcommand{\arraystretch}{1.2} 
% \resizebox{0.99\textwidth}{!}
{
\begin{tabular}{p{0.23\textwidth} p{0.75\textwidth}}
\toprule
\textbf{Subcategory}   & \textbf{Example}      \\
\midrule
\rowcolor{gray!15}
\multicolumn{2}{l}{\emph{Multi-goal tasks.}}    \\ 
Single-app related tasks & First, \underline{find out what day of the week it is today}, then open Amap and \underline{set my vehicle} as a small passenger car with the license plate Jing ABDCF0 and fuel type gasoline. After that, go to the Tools tab, find the ``Driving Restriction Inquiry" feature, and \underline{check whether there are any restrictions for today}.  \\
Single-app unrelated tasks & First, open HelloBike in \underline{Alipay} and check the bike card purchase page. Then go to Ant Insurance to view my insurance policies. Finally, go to Alipay Membership to see all my privileges.   \\
Cross-app related tasks &  {Remove classmates} Zhang San, Zhao Si, and Li Wu from your \underline{QQ contacts}. Then, \underline{on Douyin}, mark the first recommended video on the home page as ``Not Interested".  \\
Cross-app unrelated tasks & Open Trip.com and find the earliest flight from San Diego to San Francisco for tomorrow, note the actual departure time and estimated arrival time, text Mom: ``My flight is expected to \underline{arrive at [arrival time]} tomorrow. Please pick me up \underline{at [airport name] Airport."}, and create a calendar event on my phone titled ``FlightReminder" set for two hours before the flight's actual departure time.  \\
\rowcolor{gray!15}
\multicolumn{2}{l}{\emph{Multi-constraint tasks.}}    \\ 
Operation-level constraints tasks & On AliExpress, find a Xiaomi smartphone \underline{under \$200 with 8GB RAM and 256GB storage,} \underline{in black, with 33W fast charging}; pick the best-selling one and add it to my cart.   \\
Page-Level constraints tasks & I need you to open WebNovel, \underline{go to Explore, switch to Comics, and select the School category}. Filter for Translation titles with 100–500 chapters that are ongoing and updated in the last 30 days, sort by last updated, and then open the first chapter of the second comic in the results. \\
\rowcolor{gray!15}
\multicolumn{2}{l}{\emph{Domain-specific tasks.}}    \\ 
Batch operation tasks & Please use Zoom to \underline{schedule three meetings} for me: ``Stand-up meeting" tomorrow at 10:00 AM for 30 minutes, set to repeat daily; ``Weekly Meeting" next Monday at 2:00 PM for 30 minutes, set to repeat weekly; and ``Mid-year summary" the day after tomorrow at 7:00 PM for 1 hour. Use my local time. \\
Combine with VLM capabilities tasks &  Open Zhihu and search for ``What books are needed to prepare for IELTS?". Sort the results by number of upvotes and open the third article. Read through it and \underline{identify the titles of the three recommended books} mentioned there. Then open JD.com, search for these three books one by one, and add the best-selling in-stock item for each book to your shopping cart.  \\
Multi-step single-goal tasks &  Help me use the satellite rescue feature in Amap (Gaode Map) to \underline{request vehicle assistance}. Call a tow truck to tow my car to Building A of the China World Trade Center. The contact phone number is 1504042147 and the license plate number is Jing AABCD1. Then initiate the rescue request.  \\

\bottomrule
\end{tabular}}
\end{table*}

\begin{figure*}[t]
    \centering
    \includegraphics[width=\linewidth]{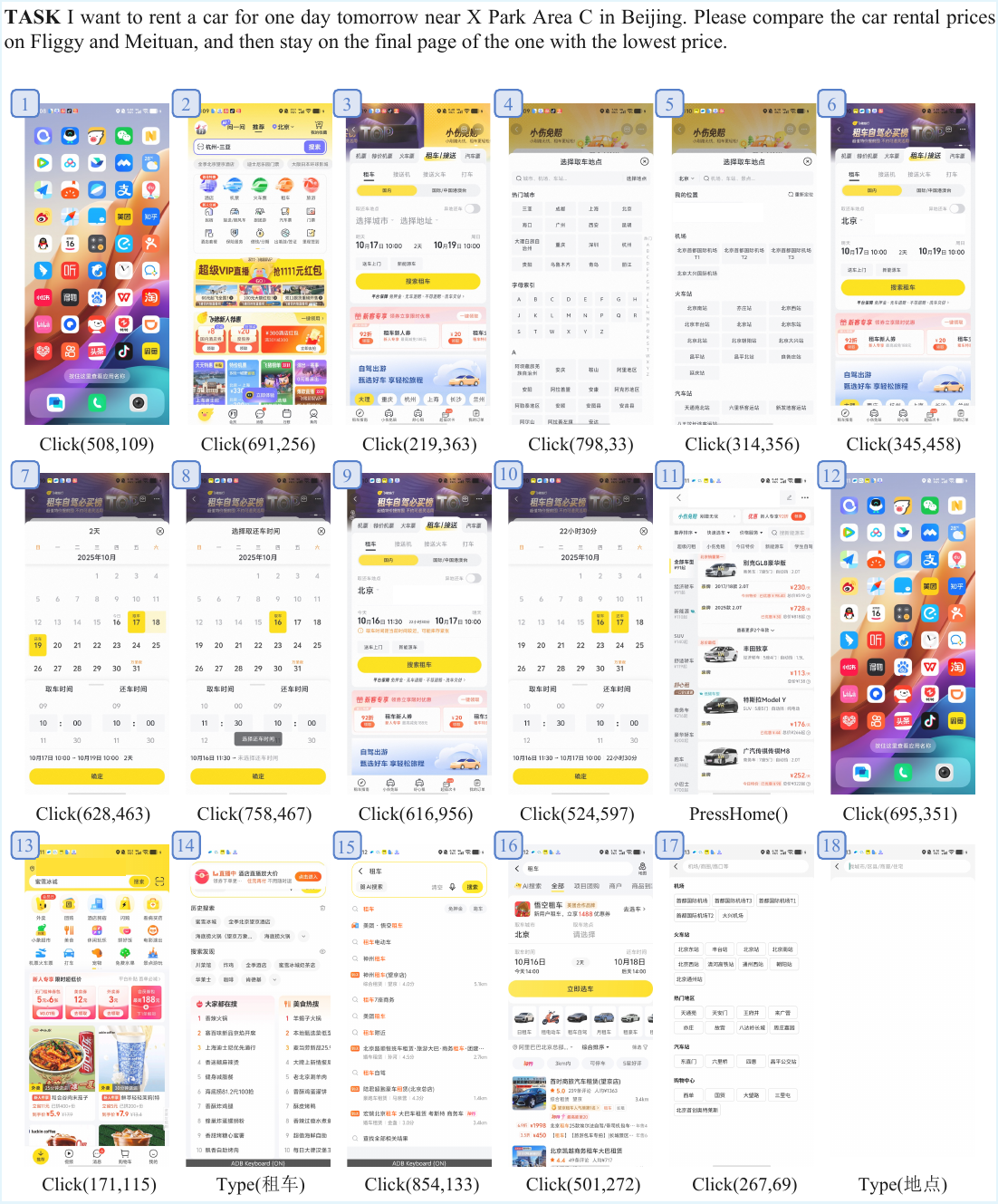}
    \caption{\textbf{Example of Annotation Trajectories for Cross-App Related Tasks.}}
    \label{fig:vis_traj_1}
    % \vspace{-0.4cm}
\end{figure*}

\begin{figure*}[t]
    \centering
    \includegraphics[width=\linewidth]{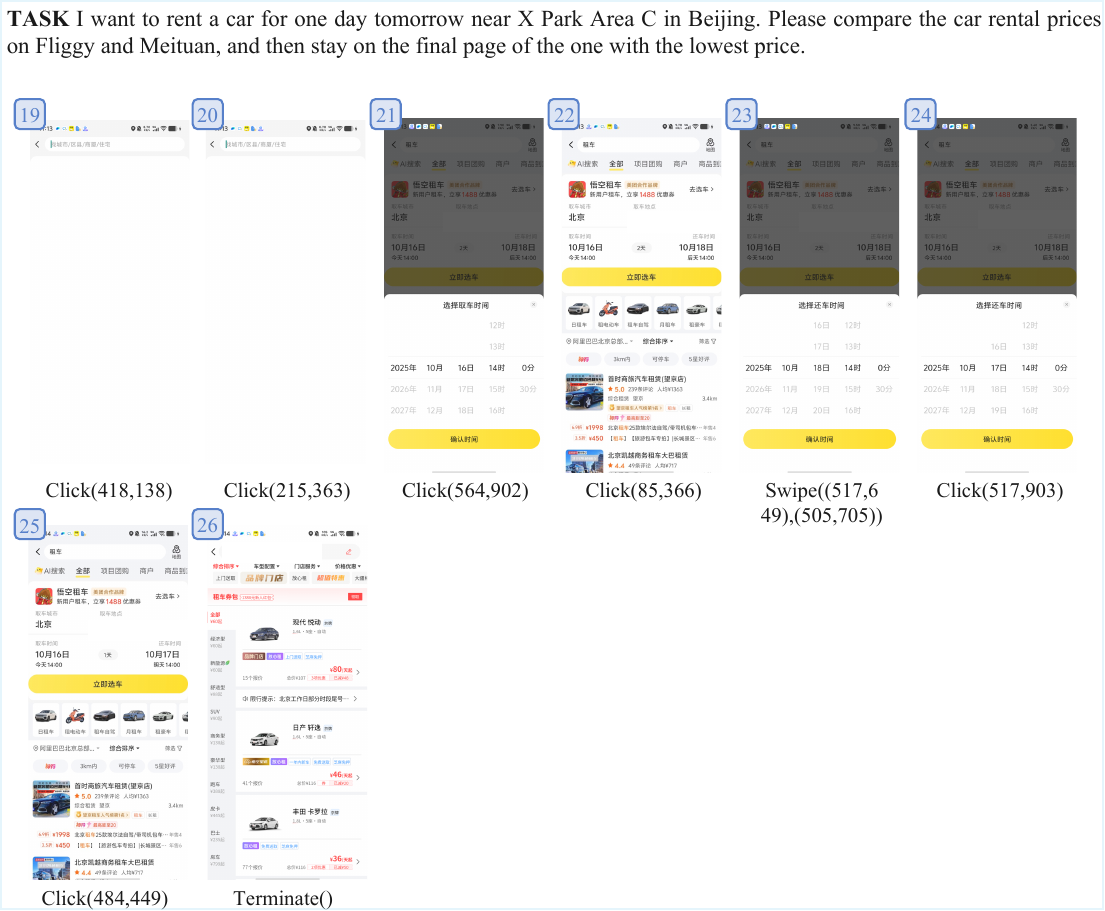}
    % \vspace{-0.4cm}
    \caption{\textbf{Example of Annotation Trajectories for Cross-App Related Tasks.}}
    \label{fig:vis_traj_2}
    % \vspace{-0.4cm}
\end{figure*}

In addition, we selected a price comparison case from the cross-app related tasks to more intuitively demonstrate the task execution process and annotation details. Specifically, as shown in \cref{fig:vis_traj_1} and \cref{fig:vis_traj_2}, we provide the complete annotation operation trajectory and the corresponding screenshots.

\section{Visualization Examples}\label{sec:ad_case}

\subsection{Agent’s Action Trajectory}
\begin{figure*}[t]
    \centering
    \includegraphics[width=0.95\linewidth]{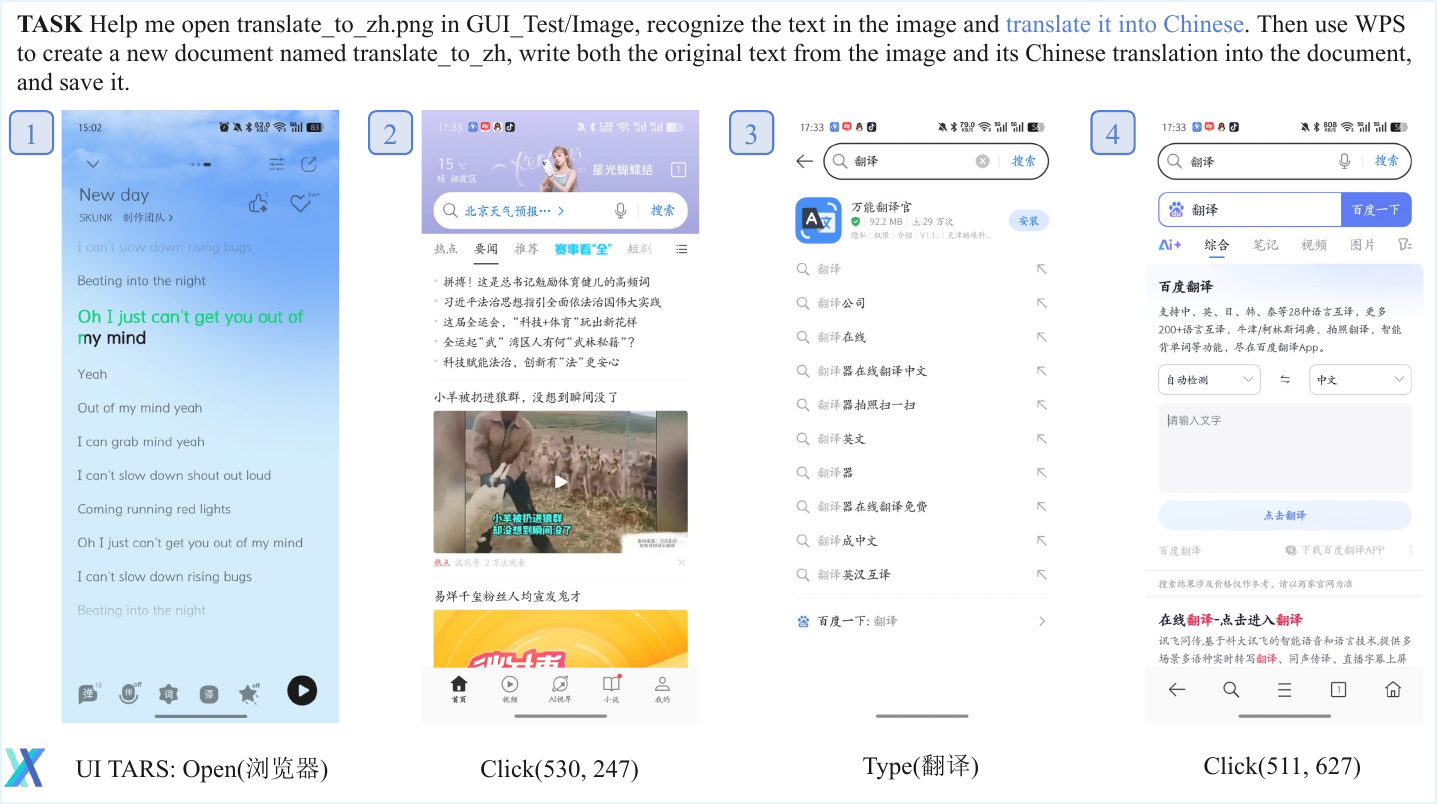}
    % \vspace{-0.4cm}
    \caption{\textbf{A Visual Example of Agent Execution Requiring the VLM’s Own Capabilities.}}
    \label{fig:vis_translate}
    % \vspace{-0.4cm}
\end{figure*}

\begin{figure*}[t]
    \centering
    \includegraphics[width=0.95\linewidth]{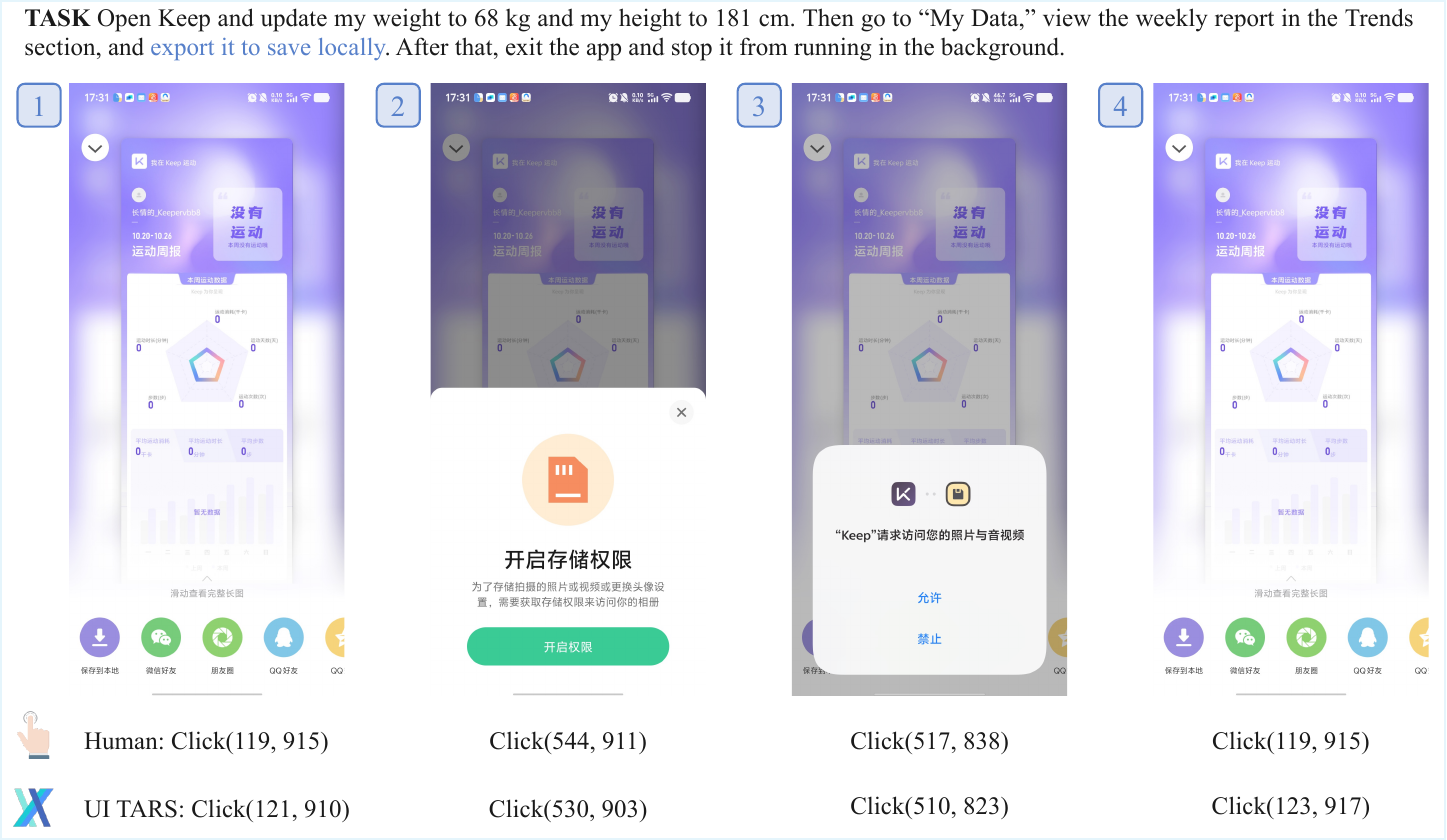}
    % \vspace{-0.4cm}
    \caption{\textbf{A Visual Example of an Agent Handling Missing Permissions.}}
    \label{fig:vis_per}
    % \vspace{-0.4cm}
\end{figure*}

To qualitatively analyze how existing models perform on \modelname, we present several examples of their actual operation processes during dynamic evaluation on \modelname. As shown in \cref{fig:vis_translate}, we illustrate how UI TARS handles a case that requires the multimodal large model’s own inherent capabilities. In this example, the task is to translate the content of an image into Chinese. Although the instruction is to translate it directly, the model takes a more complicated approach by invoking a third-party translation tool. As shown in \cref{fig:vis_per}, we demonstrate how the model responds to a missing-permission issue and is able to effectively determine and obtain the required permissions. However, this also leads us to reflect on the potential security risks introduced by such automated workflows.

\subsection{Unexpected Event Examples}
\begin{figure*}[t]
    \centering
    \includegraphics[width=0.95\linewidth]{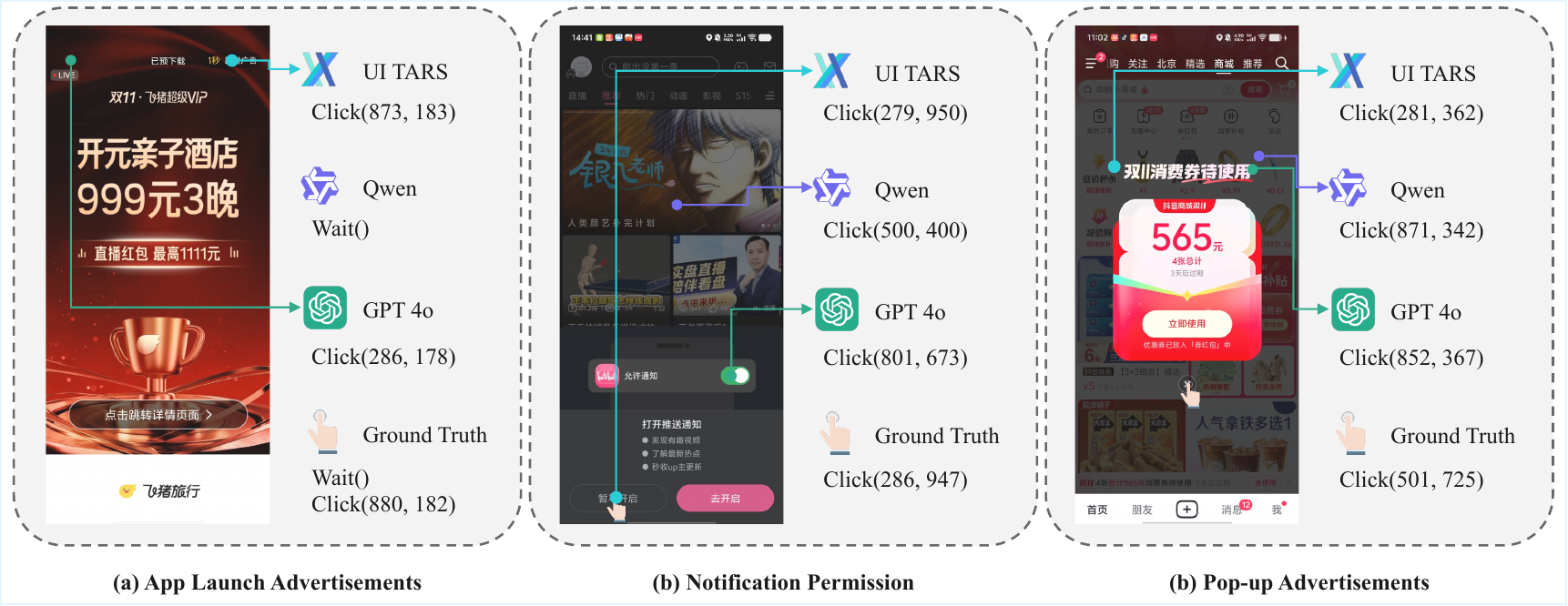}
    % \vspace{-0.4cm}
    \caption{\textbf{A Visual Example of an Agent Handling Various Emergencies.}}
    \label{fig:vis_ad}
    % \vspace{-0.4cm}
\end{figure*}

As described in the main text, we incorporated a large amount of data related to unexpected events into \modelname to evaluate the model’s reflection and self-correction capabilities. \cref{fig:vis_ad} shows the model’s performance in several representative, unanticipated scenarios. Encouragingly, the model demonstrates a certain degree of competence in handling sudden events. However, its predictions still contain some localization errors and hallucinations. For example, the agent repeatedly and incorrectly assumes that the close button is located in the top-left or top-right corner.

\end{document}